\documentclass[11pt]{article}

\newcommand{\draftonly}[1]{#1}
\renewcommand{\draftonly}[1]{}

\usepackage[preprint]{acl}

\usepackage{times}
\usepackage{latexsym}

\usepackage[T1]{fontenc}

\usepackage[utf8]{inputenc}

\usepackage{microtype}

\usepackage{graphicx}
\usepackage{linguex}
\usepackage{placeins}
\usepackage{amsmath}
\usepackage[skins,raster]{tcolorbox}

\definecolor{promptcolor}{HTML}{F7F7F7}
\definecolor{basecolor}{HTML}{FFF4D6}
\definecolor{poscolor}{HTML}{E6E6FA}
\definecolor{negcolor}{HTML}{D8F3DC}

\newcommand{\genbox}[3]{%
    \vspace{0.2em}
    \noindent\colorbox{#1}{%
    \begin{minipage}{\dimexpr\linewidth-2\fboxsep\relax}
    \raggedright
    \textbf{#2:} {\itshape #3}
    \end{minipage}}%
}

\newtcolorbox{promptbox}{
    colback=promptcolor,
    colframe=black!50,
    boxrule=0.4pt,
    arc=1pt,
    width=0.94\linewidth,
    left=6pt, right=6pt, top=6pt, bottom=6pt,
    boxsep=0pt,
}

\newtcolorbox{respbox}[2]{
    colback=#1, colframe=black!40, boxrule=0.3pt,
    title=#2, fonttitle=\bfseries\small,
    coltitle=black, colbacktitle=#1!70!white,
    fontupper=\itshape\small,
}

\usepackage{microtype}

\usepackage{amssymb,bm,mathtools,color}

\usepackage[normalem]{ulem}
\usepackage{enumitem}

\usepackage{booktabs}
\usepackage{multirow}
\usepackage{tabularx}
\usepackage{multicol}
\usepackage{multirow}
\usepackage{colortbl}
\usepackage{hhline}
\usepackage{stfloats}

\newcolumntype{Y}{>{\centering\arraybackslash}X}

\usepackage{xcolor}
\definecolor{red}{rgb}{0.74,0.08,0.10}
\definecolor{green}{rgb}{0.26,0.49,0.18}
\definecolor{blue}{rgb}{0.22,0.53,0.75}

\usepackage{colortbl}
\definecolor{Gray}{gray}{0.9}
\definecolor{LightCyan}{rgb}{0.75,1,1}

\makeatletter
\newcommand\notsotiny{\@setfontsize\notsotiny\@viiipt\@ixpt}
\makeatother
\usepackage{booktabs}
\usepackage{multirow}
\usepackage{graphicx}

\usepackage{pifont}
\usepackage{amsmath}
\usepackage{amssymb}
\usepackage{ifthen}
\usepackage{dsfont}
\usepackage{bold-extra}
\usepackage{courier}

\usepackage{listings}
\usepackage{xcolor}
\usepackage{ltablex}
\usepackage{longtable}
\usepackage{multicol}

\usepackage{tikz}
\usetikzlibrary{bayesnet}
\usepackage[linguistics]{forest}

\usepackage{caption}
\usepackage{cancel}

\usepackage{xspace}
\usepackage{comment}
\usepackage{color,soul}

\usepackage{float}

\usepackage[noabbrev,capitalize]{cleveref} 

\crefname{page}{page}{pages}
\crefname{footnote}{footnote}{footnotes}   
\crefname{equation}{equation}{equations}   

\creflabelformat{equation}{#2\textup{#1}#3}

\crefname{line}{line}{lines}               
\crefrangeformat{line}{lines #3#1#4--#5#2#6}
\crefname{lstlsting}{Listing}{Listings}   
\crefname{section}{\S}{\S\S}
\Crefname{section}{\S}{\S\S}    
\crefformat{section}{#2\S#1#3}  
\Crefformat{section}{#2\S#1#3}
\crefrangeformat{section}{\S\S#3#1#4--#5#2#6}
\Crefrangeformat{section}{\S\S#3#1#4--#5#2#6}
\crefmultiformat{section}{\S#2#1#3}{ and~\S#2#1#3}{, \S#2#1#3}{ and~\S#2#1#3}
\Crefmultiformat{section}{\S#2#1#3}{ and~\S#2#1#3}{, \S#2#1#3}{ and~\S#2#1#3}
\crefrangemultiformat{section}{\S\S#3#1#4--#5#2#6}{ and~\S\S#3#1#4--#5#2#6}{, \S\S#3#1#4--#5#2#6}{ and~\S\S#3#1#4--#5#2#6}
\Crefrangemultiformat{section}{\S\S#3#1#4--#5#2#6}{ and~\S\S#3#1#4--#5#2#6}{, \S\S#3#1#4--#5#2#6}{ and~\S\S#3#1#4--#5#2#6}

\newcommand{\prob}[2][]{p\ifthenelse{\not\equal{}{#1}}{_{#1}}{}(#2)} 
\newcommand{\expect}[2][]{\text{\bf E}\ifthenelse{\not\equal{}{#1}}{_{#1}}{}\!\left[#2\right]}
\newcommand{\var}[2][]{\text{\bf Var}\ifthenelse{\not\equal{}{#1}}{_{#1}}{}\!\left[#2\right]}

\newcommand{\ie}{\emph{i.e.}, }

\newcommand{\makename}[3][s]{%
  \expandafter\newcommand\csname #2\endcsname{#3\xspace}%
  \expandafter\newcommand\csname #2s\endcsname{#3#1\xspace}%
}

\makename{aucmse}{\texttt{AUC-MSE}}
\makename{scauc}{\texttt{SCAUC}}

\usepackage[]{todonotes}

\usepackage[]{todonotes}  
\makeatletter
\newcommand*\iftodonotes{\if@todonotes@disabled\expandafter\@secondoftwo\else\expandafter\@firstoftwo\fi}  
\makeatother

\usepackage{inconsolata}

\newcommand{\ambient}{\textsc{AmbiEnt}}

\newcommand{\setamb}{\mathcal{S}_{\mathrm{amb}}}
\newcommand{\setclear}{\mathcal{S}_{\mathrm{clear}}}

\newcommand{\scoreamb}{S_{\mathrm{amb}}}
\newcommand{\scoreclear}{S_{\mathrm{clear}}}
\newcommand{\scoresel}{\operatorname{Sel}}

\newcommand{\cmulogo}[2]{%
  \raisebox{#1}{\includegraphics[height=0.8em]{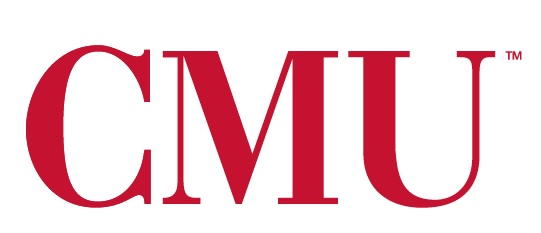}}%
  \hspace{0.1em}%
  \raisebox{#2}{\includegraphics[height=0.61em]{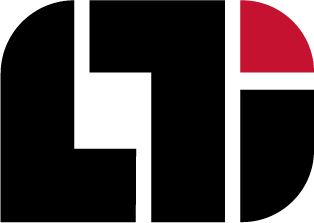}}%
}

\title{Improving Confidence Estimation in Language Models with Linear Representations of Ambiguity}

\author{%
    Themistoklis Nikas\,%
    \raisebox{0.3ex}{\includegraphics[height=0.8em]{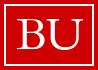}}
    \qquad
    Nishant Subramani \cmulogo{0.3ex}{0.45ex}
    \qquad
    Aaron Mueller\,%
    \raisebox{0.3ex}{\includegraphics[height=0.8em]{figures/BU_logo.png}}
    \\
    \normalfont
    \raisebox{-0.2ex}{\includegraphics[height=0.8em]{figures/BU_logo.png}}:
    Boston University
    \\
    \normalfont
    \cmulogo{-0.2ex}{-0.05ex}:
    Carnegie Mellon University, Language Technologies Institute
    \\
    \normalfont \texttt{tnikas@bu.edu}
}

\begin{document}
\maketitle

\frenchspacing




\begin{abstract}
As language models (LMs) are increasingly deployed in downstream applications, accurately estimating their likelihood of success on a given example is critical. However, confidence estimates from LMs are poor predictors of their likelihood of success. Accurate confidence estimation is linked in part to the detection of \emph{aleatoric uncertainty}; we hypothesize that LMs possess representations of at least some aspects of this uncertainty in the form of representations of the inherent ambiguity of the input. Using difference-in-means on minimal pairs of ambiguous and unambiguous sentences, we derive ambiguity representations in LM activations, and find that steering with them selectively (albeit weakly) modulates LMs' judgments of the input's ambiguity. We then
propose DART, a fine-tuning method that encourages updates to lie within the span of the ambiguity directions, and penalizes changes to the LM activations outside of this subspace. DART greatly improves calibration, encodes these improvements into the model parameters, and maintains model accuracy on multiple-choice and generative QA tasks.
\end{abstract}

\section{Introduction}
While language models (LMs) have demonstrated impressive capabilities, assessing their trustworthiness remains challenging. LMs do not achieve uniformly good performance on all tasks; as they are increasingly deployed in high-stakes applications, it is increasingly important to be able to accurately estimate their likelihood of success on a given input \citep{Dawid1982TheWBA,naeini2015obtaining,pmlr-v70-guo17a}. \emph{Confidence estimation} entails predicting the likelihood that a system's predicted answer is correct; a confidence estimator is said to be \emph{well calibrated} if on an unseen test distribution, the system is correct approximately as often as its confidence estimate predicts it should be.
Well-calibrated probabilities can guide downstream decision making by allowing a model to, for example, request clarification or defer to a search.
Despite this need, LMs are known to be poor estimators of their own likelihood of success \citep{desai-durrett-2020-calibration, jiang-etal-2021-know, stengel-eskin-van-durme-2023-calibrated}.

Recent interpretability findings suggest that LMs internally represent signals relevant to evaluating their own answers. For example, prior work has identified representations associated with the truthfulness of a model's generated answer to a query \citep{marks2024geometrytruthemergentlinear,orgad2025llms}, and model representations have been used to improve confidence estimation~\citep{subramani-etal-2025-mice, SubramaniAcute2026}. 
The existence of these representations does not guarantee that a model can actually express well-calibrated confidence.
It does, however, suggest an opportunity: representations identified via interpretability methods may provide useful structure for \emph{adapting} a model's behavior. In this study, we ask: can we use representations of a relevant linguistic property to guide fine-tuning toward better-calibrated predictions?

We investigate ambiguity as one such property. When an input can be interpreted multiple ways, a model may confidently answer under one interpretation while overlooking alternatives. For example, ``Where is the bank?'' could refer to a financial institution or the edge of a river. Without clarification, confidence in a particular response should account for this irreducible uncertainty in the input (what has been called \emph{aleatoric uncertainty}).\footnote{We note that this is not a complete picture of what should make a model uncertain: there is also epistemic uncertainty, which relates to a specific model's knowledge regarding a subject. We focus on aleatoric uncertainty here, as the same data could be used for any given model, whereas epistemic uncertainty is highly model-dependent.} We ask whether representations derived via paired examples that are ambiguous or not can provide a useful inductive bias for learning from general task data.

Using the \textsc{AmbiEnt} dataset \citep{liu2023wereafraidlanguagemodels} and the ambiguity taxonomy of \citet{li2024taxonomyambiguitytypesnlp}, we extract linear representations that modulate LM judgments of input ambiguity via difference-in-means \citep{marks2024geometrytruthemergentlinear} on minimal pairs of ambiguous and disambiguated sentences. We show that the causal influence of these directions is selective for ambiguous sentences in the lower-middle layers of the model: steering with them shifts the model's judgment of an input's ambiguity more on ambiguous inputs than disambiguated ones. Inference-time steering along these directions sometimes improves calibration for multiple language models from different families on the MMLU and TruthfulQA datasets, although improvements are sporadic.

To more reliably encode these benefits into the model's parameters and remove the need for inference-time interventions, we propose \textbf{D}irection-\textbf{A}ligned \textbf{R}egularization for \textbf{T}uning (DART), a fine-tuning procedure that regularizes changes in LM activations outside the subspace spanned by a set of steering vectors. Across models of different families and sizes, DART consistently and significantly improves confidence estimation; its improvements are larger than those obtained via vanilla LoRA, over DART using a random subspace of the same rank, and relative to inference-time steering. It also has little to no impact on task accuracy.

\begin{figure*}[t]
    \centering
    \includegraphics[width=\textwidth]{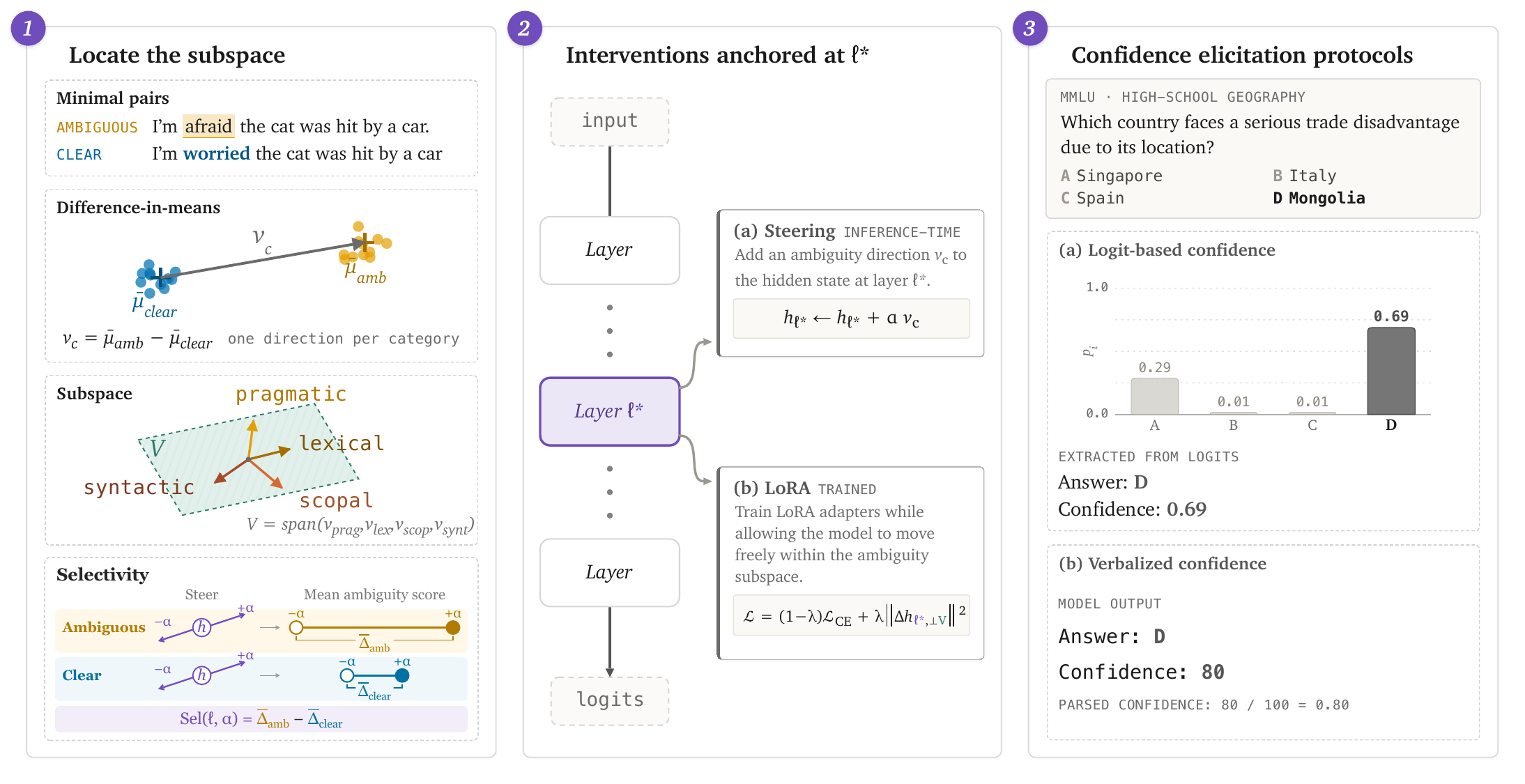}
    \caption{
    \textbf{Overview.} We locate representations of ambiguity in language models using difference-in-means (\cref{sec: discovering_ambiguity_directions}), and find that these qualitatively modulate the model's confidence in its reasoning (\cref{sec: inference_time_steering_conf_est}). Quantitatively, steering into ambiguity directions tends to improve confidence calibration. We propose a LoRA-based procedure which regularizes updates orthogonal to these ambiguity representations whereby improving calibration (\cref{sec: results}).
    }
    \label{fig:pipeline}
\end{figure*}

In summary, our contributions are:\footnote{We will release our data and code upon deanonymization.}
\begin{itemize}[topsep=0pt,noitemsep]
    \item We propose a procedure to find representations that modulate LM judgments of input ambiguity, and validate the behavioral influence of these vectors on models' judgments of ambiguity.
    \item We introduce DART, which uses these representations to regularize fine-tuning. This encodes calibration improvements into the model's parameters (primarily by mitigating overconfidence), often yields further confidence estimation improvements over baselines, and maintains accuracy on QA tasks.
\end{itemize}

\section{Related Work}

\paragraph{Confidence estimation \& calibration.} 
\label{par:post-hoc-calibration}
Confidence estimation is well-studied in both the machine learning and natural language processing (NLP) communities. 
Most methods operate as post-hoc calibration methods, operating on a single confidence estimate and recalibrating that in various ways including histogram regression~\citep{nobel1996histogram}, Nadaraya-Watson kernel regression~\citep{nadaraya1964estimating, watson1964smooth}, Platt scaling~\citep{platt1999probabilistic}, isotonic regression~\citep{barlow1972statistical}, and beta calibration~\citep{pmlr-v54-kull17a}.
There exist multiple methods to estimate the confidence of a language model output: taking the joint probability of the sequence under the language model~\citep{desai-durrett-2020-calibration}, asking the model directly via a prompt (linguistic calibration; ~\citealp{mielke-etal-2022-reducing, tian-etal-2023-just, band2024linguistic}), and using model internals~\citep{liu-etal-2024-enhancing-language, subramani-etal-2025-mice, SubramaniAcute2026, kumaran2026llmsdetectcorrecterrors}.

\paragraph{Ambiguity in NLP.} A well-established line of work treats ambiguity in downstream tasks as a property of the inputs, while others treat it as an artifact of annotation noise. \citet{pavlick-kwiatkowski-2019-inherent} show that disagreements in NLI reflect valid variation in interpretations of an input, and \citet{plank-2022-problem} argues that this label variation should be directly modeled rather than marginalized away. This has implications for confidence estimation: \citet{baan-etal-2022-stop} show standard confidence estimation metrics break when inputs admit multiple interpretations. Recent benchmarks have evaluated model sensitivity to specific ambiguities, such as ambiguity in open-domain question answering \citep{min-etal-2020-ambigqa}, scope ambiguities \citep{kamath-etal-2024-scope}, among others \citep{liu2023wereafraidlanguagemodels,li2024taxonomyambiguitytypesnlp}, with mixed results. We do not contribute to ambiguity modeling in this work, nor to resolving it; rather, we use controlled manipulations of input ambiguity to measure whether there exist representations that track known ambiguities in inputs, and whether these representations modulate a model's estimation of its own confidence.

\paragraph{Actionable interpretability.} Early interpretability research on pretrained models (often termed ``BERTology'') proposed methods for locating \citep[\emph{inter alia}]{hewitt-manning-2019-structural, jawahar-etal-2019-bert,tenney-etal-2019-bert, niu-etal-2022-bert, li2025model, acevedo2026differential} and controlling \citep{giulianelli-etal-2018-hood} language model representations sensitive to linguistic structure. Similar work has recently emerged from the (mechanistic) interpretability literature; this includes steering vectors \citep{subramani-etal-2022-extracting}, which allow one to control properties of the outputs of an LM at inference time. The most common method for locating steering vectors is difference-in-means \citep{marks2024geometrytruthemergentlinear,rimsky-etal-2024-steering}; difference-in-means has been shown to be empirically effective for monitoring and controlling a wide variety of concepts in language models \citep{wu2025axbench}.

Our work aligns with a growing call for \emph{actionable} or \emph{pragmatic} interpretability, which argues that interpretability research should be evaluated by its practical downstream utility rather than its explanatory power alone \citep{nanda2025pragmatic,orgad2026interpretabilityactionable}. Our work is aligned with this perspective: we demonstrate practical benefits to isolating representations of input ambiguity. Recent work in this line has largely targeted behavioral changes, such as refusal \citep{arditi2024refusal}, sycophancy \citep{chen2025personavectorsmonitoringcontrolling}, or hallucinations \citep{prasad2026featuresrewardsscalablesupervision}. We instead target confidence calibration, and show that a small set of directions can be used to directly improve it.

\section{Discovering Ambiguity Directions}
\label{sec: discovering_ambiguity_directions}

We locate directions in language model activations that distinguish ambiguous sentences from disambiguated rewrites of those sentences. Our method uses paired sentences 
$(x_i^{\mathrm{amb}}, x_i^{\mathrm{clear}})$, where
$x_i^{\mathrm{amb}}$ is an ambiguous sentence, and $x_i^{\mathrm{clear}}$ is a rewrite that resolves the ambiguity.

\subsection{Data}
We use \ambient{} \citep{liu2023wereafraidlanguagemodels} to source sentence pairs. \ambient{} provides minimal pairs of ambiguous sentences together with disambiguated rewrites. For sentences that have multiple disambiguations, we create one paired instance for each matched rewrite.

There exist different types of ambiguity, including scopal, pragmatic, lexical, and syntactic ambiguities (among others). These ambiguity types could have distinct representations; thus, we split the data $\mathcal{D}$ by ambiguity category $c$, resulting in category-specific datasets $\mathcal{D}_c$. Using the annotations of \citet{li2024taxonomyambiguitytypesnlp}, we categorize each example into an ambiguity category. See App.~\ref{app:Ambient} for dataset details and examples.

We restrict direction discovery to the four largest ambiguity categories: \textit{pragmatic, lexical, scopal} and \textit{syntactic}.\footnote{The remaining categories have fewer than 15 validation examples, which is too few for reliable evaluation of candidate directions in the selection procedure described in~\cref{sec:intervening-amb-directions}. Category counts are reported in~App.~\ref{app:Ambient},~\cref{tab:category-train-val-counts}.} 
For each category, we split examples into 80\% train and 20\% validation sets. If multiple disambiguated sentences were derived from the same ambiguous example, we ensure that each derived example is placed in the same split. 

\subsection{Locating Ambiguity Directions}
To identify directions that encode ambiguity, we follow prior work in computing a difference between the mean activation vectors of paired inputs \citep{marks2024geometrytruthemergentlinear, rimsky-etal-2024-steering}. Given ambiguity category $c$
and paired examples $(x_\text{amb}, x_\text{clear})$ in $\mathcal{D}_c$, we format each sentence with the same ambiguity aware prompt and extract hidden activations $\mathbf{h}_{\ell}(x) \in \mathbb{R}^d$ at each transformer layer $\ell \in \{1,\ldots,L\}$.\footnote{See~App.~\ref{app:prompts} for prompt and tokenization details.}
For each category $c$ and layer $\ell$ 
we define the candidate ambiguity vector $\mathbf{v}_{c,\ell} \in \mathbb{R}^d$ as the mean difference: 
\begin{equation}
\mathbf{v}_{c,\ell}
=
\frac{1}{N_c}
\sum_{i=1}^{N_c}
\left(
\mathbf{h}_{\ell}(x_{\mathrm{amb}}^{(i)})
-
\mathbf{h}_{\ell}(x_{\mathrm{clear}}^{(i)})
\right),
\end{equation}
where $N_c$ is the number of training pairs for each category $c$. This produces a set of $L$
candidate vectors for each category. Henceforth, we omit the category index $c$ for clarity when describing computations applied to each ambiguity category.

\subsection{Intervening with Ambiguity Directions}
\label{sec:intervening-amb-directions}
A raw intervention scale is not directly comparable across layers, tokens or models, since hidden state norms vary with depth and architecture. We therefore express steering strength as a percentage of a reference hidden-state norm computed separately for each layer.
Given a steering percentage 
$
\alpha \in \mathcal{A}
=
\{0.1\%,\allowbreak
0.2\%,\allowbreak
0.5\%,\allowbreak
1\%,\allowbreak
3\%,\allowbreak
5\%,\allowbreak
10\%,\allowbreak
50\%,\allowbreak
100\%,\allowbreak
150\%,\allowbreak
200\%\},
$\footnote{For example, $\alpha=10\%$ means that the ambiguity vector is added at $10\%$ of the activation vector's norm at layer $\ell$.} we intervene by adding a perturbation along this direction at the same layer:
\begin{equation}
\tilde{\mathbf{h}}_{\ell}(x)
=
\mathbf{h}_{\ell}(x)
+
\alpha \rho_{\ell}\frac{\mathbf{v}_{\ell}}{\lVert \mathbf{v}_{\ell} \rVert}.
\end{equation}
Here, $\rho_\ell$ refers to the average norm of the activation vector $\mathbf{h}_\ell$ on a given distribution. Thus, this can be interpreted as shifting the activation vector by a given percentage $\alpha$ toward the ambiguity direction.\footnote{We use this percentage formulation rather than using $\alpha$ as a multiplicative constant on $\mathbf{v}_\ell$ directly because it yields more stable and comparable hyperparameter settings across language models with very different activation norms.} By construction, positive steering moves the representation in the ambiguous direction, whereas negative steering moves it in the clear direction. The modified hidden state is then propagated through the remaining layers of the model.

\paragraph{Evaluating selectivity.} For our purposes, a useful ambiguity direction should not necessarily induce a uniform behavioral change in the model across examples; rather, it should be \emph{selective} for ambiguous inputs. That is, steering should have a stronger effect on ambiguous inputs than on their clear paired examples, such that interventions along the discovered direction should result in better-calibrated sensitivity to an input's ambiguity. To evaluate the selectivity of an ambiguity direction, we compare its intervention effect on ambiguous inputs with its effect on their clear counterparts. For a sentence $s$, we first define a model's ambiguity preference as 
\begin{equation}
\psi(s)
=
\log P(\mathrm{Yes}\mid s)
-
\log P(\mathrm{No}\mid s),
\end{equation}
where higher values indicate a stronger preference for an ambiguous interpretation under an ambiguity judgment prompt (defined in App.~\ref{app:prompts}). For an ambiguity direction $\hat{\mathbf{v}}_{\ell}$ and steering percentage 
$
\alpha
\in
\mathcal{A}
$
, we define the sentence level steering effect:
\begin{equation}
\Delta_{\ell}(s;\alpha)
= 
\psi_{\ell}(s;+\alpha) 
- 
\psi_{\ell}(s;-\alpha)
\end{equation}
where $\psi_{\ell}(s;+\alpha)$ and $\psi_{\ell}(s;-\alpha)$ denote the ambiguity score under positive and negative interventions respectively, at steering percentage $\alpha$. We then average the steering effects over both ambiguous and clear sentences separately:

\begin{align}
\scoreamb(\ell,\alpha)
&=
\frac{1}{|\setamb|}\sum_{s\in\setamb}\Delta_{\ell}(s;\alpha) \\
\scoreclear(\ell,\alpha)
&=
\frac{1}{|\setclear|}\sum_{s\in\setclear}\Delta_{\ell}(s;\alpha)
\end{align}

Finally, we define selectivity at steering percentage $\alpha$ as the difference in mean steering effects between ambiguous and clear sentences:
\begin{equation}
\scoresel(\ell,\alpha)
=
\scoreamb(\ell,\alpha)
-
\scoreclear(\ell,\alpha).
\end{equation}
Higher selectivity indicates that the intervention shifts ambiguity scores more strongly for ambiguous inputs than for paired clear inputs.

To select the optimal layer $\ell^\ast$ for steering, we score each candidate vector at each steering percentage 
$
\alpha
\in
\mathcal{A}
$
separately for each ambiguity category. For downstream experiments, we select the layer that maximizes selectivity for each model. We first average selectivity over $\mathcal{A}$ for each category $c$:
\begin{equation}
\overline{\scoresel}_c(\ell)
=
\frac{1}{|\mathcal{A}|}
\sum_{\alpha \in \mathcal{A}}
\scoresel_c(\ell,\alpha).
\end{equation}
We then aggregate across categories by weighting each category according to its number of validation pairs, to obtain $\overline{\scoresel}_{w}(\ell)$.
Finally, we select the shared layer with the highest weighted selectivity:
\begin{equation}
\ell^\ast
=
\operatorname*{arg\,max}_{\ell}
\overline{\scoresel}_{w}(\ell).
\end{equation}

Intuitively, this procedure selects the layer most likely to have the highest selectivity at many steering percentages. The selected layer and the corresponding vectors 
$\{\hat{\mathbf{v}}_{c,\ell^\ast}: c \in \mathcal{C}\}$ 
are then used for all downstream experiments for the given model.

\begin{figure}[t]
    \centering
    \includegraphics[width=\linewidth]{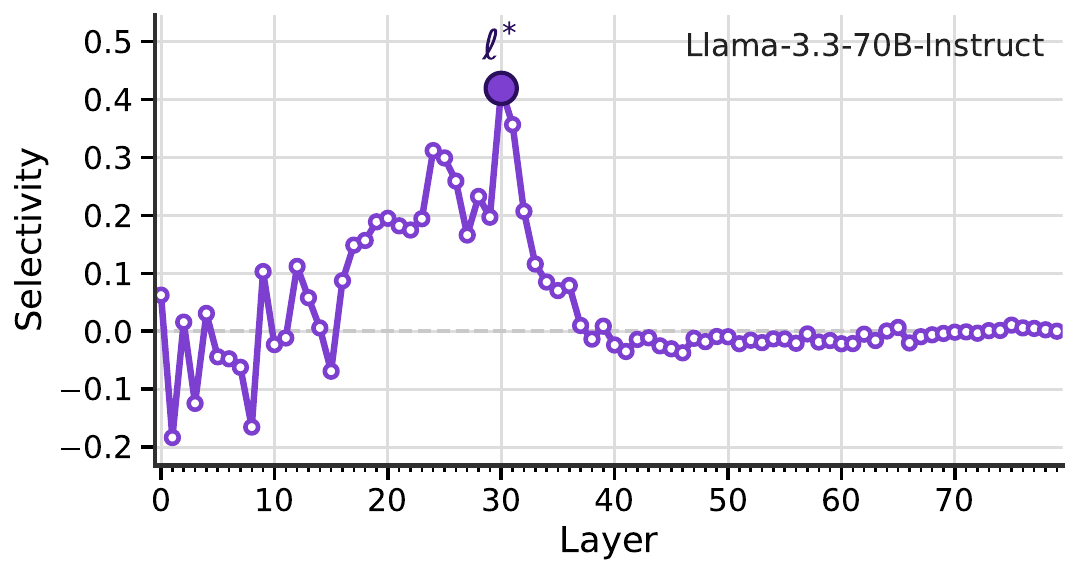}
    \caption{
    Selectivity of candidate ambiguity directions across layers for \texttt{Llama-3.3-70B-Instruct}. We show weighted mean
    selectivity averaged across relative strengths and core ambiguity
    categories. The selected layer is the highest-selectivity point.
    }
    \label{fig:selectivity-70B}
\end{figure}

Reference hidden-state norms $\rho_{\ell}$ are computed separately for
each layer,\footnote{$\rho_\ell$ is computed as the mean norm of the activation vectors for a given task dataset at layer $\ell$.} ensuring that the reported
percentage corresponds to the same relative intervention size across
layers and models, as norms are known to grow with greater depth and more pretraining~\citep{merrill-etal-2021-effects}.
\cref{fig:selectivity-70B} shows how selectivity varies across layers of \texttt{Llama-3.3-70B-Instruct}. Across all models, the optimal layer ($\ell^*$) is generally one of the lower-middle layers (App.~\ref{app:model-offset-selectivity}).

\subsection{Qualitative Inspection}
We first qualitatively observe to what extent activation steering with ambiguity vectors modulates models' expressed uncertainties.
In~\cref{tab:dalai-lama-qualitative}, we show an example from MMLU. Before steering, the model states that the question is unambiguous.
Steering towards higher ambiguity ($+\mathbf{v}_{c,\ell}$) at inference-time  leads to an output generation stating that there is some ambiguity, despite that the question clarifies that the subject is the current Dalai Lama.
When steering towards clear ($-\mathbf{v}_{c,\ell}$), the model generates text stating that there is no ambiguity (with respect to which Dalai Lama is being referenced).\footnote{Note that we make no judgments regarding the quality of these outputs nor whether the vectors we have derived are \emph{the} representations of ambiguity; for the purpose of confidence estimation, we are only interested in whether model judgments of ambiguity are affected by interventions using these steering vectors.} We observe similar trends in other examples (App.~\ref{app:additional-qual-eval}).

\begin{figure}[t]
\centering
\setlength{\fboxsep}{4pt}

\begin{minipage}{\linewidth}

\noindent\colorbox{promptcolor}{%
\begin{minipage}{\dimexpr\linewidth-2\fboxsep\relax}
\raggedright
\textbf{Prompt:}
\textup{[\ldots]} When was the \emph{current Dalai Lama} born?

A. 1935 \quad B. 2000 \quad C. 1965 \quad D. 1975.

\vspace{0.1em}
Briefly explain your interpretation and reasoning. If the question seems ambiguous or underspecified, mention why.
\textup{[\ldots]}
\end{minipage}}

\genbox{basecolor}{No intervention}{%
The question asks for the birth year of the ``current Dalai Lama.'' \textup{[\ldots]}
There is \textbf{no ambiguity} regarding the birth year of the 14th Dalai
Lama in historical records.
}

\genbox{poscolor}{Adding ambiguity vector}{%
The question asks for the birth year of the ``current Dalai Lama.'' This
phrasing can be \textbf{slightly ambiguous} because the title ``Dalai Lama''
has been held by different individuals at different times. \textup{[\ldots]}
}

\genbox{negcolor}{Subtracting ambiguity vector}{%
The question asks for the birth year of the ``current Dalai Lama.'' \textup{[\ldots]}
The question is \textbf{not ambiguous} regarding the identity of the person.
}

\end{minipage}

\caption{Qualitative evaluation of the syntactic ambiguity vector on an MMLU example. Steering positively, makes the model sensitive to the \emph{current} Dalai Lama, while subtracting it leads to a single interpretation. Additional results across prompts and datasets in~App.~\ref{app:additional-qual-eval}.}
\label{tab:dalai-lama-qualitative}

\end{figure}

\section{Using Ambiguity Directions to Improve Confidence Estimation}
\label{sec: inference_time_steering_conf_est} 
A good confidence estimator should assign higher confidence to clear inputs and lower confidence to noisy or ambiguous inputs. Having found representations that appear to modulate LM ambiguity judgments, we now 
investigate to what extent these representations can help us improve confidence estimations from language models.

\subsection{Experimental Setup}
We evaluate whether the discovered ambiguity directions improve downstream confidence estimation while preserving task performance. Our evaluation consists of multiple-choice question answering across all MMLU subjects \citep{hendryckstest2021} and TruthfulQA \citep{lin2021truthfulqa}, and open-ended generation on AmbigQA \citep{min-etal-2020-ambigqa}. For each model, method and confidence elicitation protocol, we select hyperparameters to maximize AUC-EURO on data disjoint from the evaluation set. We provide further details on the evaluation protocol in~App.~\ref{app:evaluation}.

Our experiments use \texttt{gemma-4-E4B-it} \citep{gemma4}, \texttt{Qwen3.5-4B} and \texttt{Qwen3.5-27B} \citep{qwen35}, \texttt{Llama-3.1-8B-Instruct} \citep{grattafiori2024llama3herdmodels}, \texttt{Llama-3.3-70B-Instruct} \citep{meta2024llama33}, \texttt{OLMo-3-7B-Instruct} \citep{olmo2026olmo3}, and \texttt{Phi-4-mini-instruct} \citep{abouelenin2025phi4mini}. All inputs are formatted with the instruction-tuned models' respective chat templates.
We focus experiments in the main paper on \texttt{Llama-3.3-70B-Instruct}, and present results for other models in App.~\ref{app:model-offset-selectivity},\ref{app:detailed_results},\ref{app:cross-model-results-ts}. Trends are generally consistent across models for each experiment.

\subsection{Confidence Elicitation Protocols}
\label{sec:confidence}
Confidence in LM predictions can be elicited in several ways. We consider two protocols: a \emph{logit-based} protocol, which derives confidence from the model's probability distribution over the candidate answers, and a \emph{verbalized} protocol, where the model is instructed to generate a number reflecting its confidence~\citep{tian-etal-2023-just}. Prior work has found that the relative performance of these protocols varies across models and tasks~\citep{shrivastava2023llamasknowgptsdont,tian-etal-2023-just,band2024linguistic,liu2025llm,kim2026answerdifferentconfidenceprotocol}.

In preliminary experiments, we find logit-based confidence estimation to yield consistently more well-calibrated estimates (see App.~\ref{app:detailed_results}); we therefore focus on logit-based confidence in our main evaluations. Since our setting consists of multiple-choice question answering (MCQA) with a fixed candidate set, normalized probabilities over the answers provides a natural confidence estimate for the model's prediction, following prior work on calibration in MCQA~\citep{jiang-etal-2021-know}.

In the logits protocol, the model is evaluated at the answer position; we restrict the next token distribution to the four multiple-choice labels 
$\mathcal{Y}=\{\mathrm{A},\mathrm{B},\mathrm{C},\mathrm{D}\}$.
The predicted answer is 
$
\hat{y}
=
\operatorname*{arg\,max}_{y\in\mathcal{Y}}
p(y\mid x)
$, and confidence is
$
\tilde{p}(\hat{y}\mid x)
$ 
under the renormalized probability distribution $\tilde{p}$ consisting only of answer-related tokens. For open-ended generation, we instead use the length-normalized likelihood of the generated answer as confidence (see~App.~\ref{app:evaluation}).

In the verbalized confidence protocol, following~\citep{tian-etal-2023-just}, the model generates both an answer and a numeric confidence estimate $\in [0,100]$. We parse the answer label and confidence estimate from the generated output and divide the confidence estimate by 100 such that it lies in the same range as in the logits-based procedure.
To obtain conservative estimates, generations that do not contain both a valid answer and a valid confidence are treated as a worst case scenario, \ie as incorrect predictions with maximum confidence. Full prompt templates are given in~App.~\ref{app:prompts}.

\subsection{Metrics}
To measure confidence estimation quality, we use smooth expected calibration error (smECE; \citealp{błasiok2023smootheceprincipledreliability})
and the area under the expected utility renormalized by the oracle (EURO) curve~\citep{SubramaniAcute2026}. Traditional ECE is hyperparameter dependent, which makes comparisons across methods challenging. SmECE resolves this drawback by being hyperparameter-free and is a strictly better measure of calibration error~\citep{błasiok2023smootheceprincipledreliability}.
AUC-EURO balances calibration with decision-making utility and thus is a better metric to adjudicate trust.
We also measure accuracy and area under the receiver operating characteristic curve (\citealp[AUROC;][]{swets_1973_roc}) to measure how often each model is correct.

\begin{figure*}[t]
    \centering
    \includegraphics[width=\textwidth]{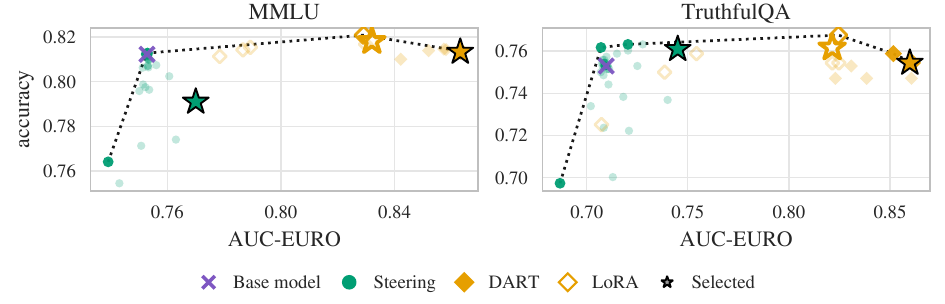}
    \caption{\textbf{Trade-off between AUC-EURO and accuracy under logit-based confidence elicitation on MMLU (left) and TruthfulQA (right)}. Results shown for \texttt{Llama-3.3-70B-Instruct}. Each point represents one hyperparameter configuration. Stars mark the selected configuration for each method. Higher values are better on both axes.
    }
    \label{fig:pareto_logits_70B}
\end{figure*}

\subsection{Methods}
\paragraph{Steering.} To perform inference-time steering on multiple choice questions, we steer at the most selective layer $\ell^\ast$, applying a positive steering intervention to the hidden states at all token positions from the end-of-turn token through the final prompt token. We try steering using ambiguity category--specific vectors, as well as a vector derived by taking the mean of all category unit vectors.

\paragraph{DART.} To permanently encode ambiguity-related updates into the parameters of a language model, we propose DART, a fine-tuning method that encourages updates to lie within the span of a set of directions (\emph{e.g.} ambiguity vectors).

We use the discovered directions to define a regularization objective for LoRA fine-tuning~\citep{hu2021loralowrankadaptationlarge}.\footnote{Although we use LoRA, DART can be applied to any objective and is compatible with any tuning method.} Let $\hat{\mathbf{v}}_1,\ldots,\hat{\mathbf{v}}_k$ denote the normalized core ambiguity directions at the selected layer $\ell^\ast$ and let
$
\mathbf{V} 
= 
[\hat{\mathbf{v}}_1,\ldots,\hat{\mathbf{v}}_k]
$
be the matrix having as columns these directions. The corresponding ambiguity space is $\operatorname{span}\{\hat{\mathbf{v}}_1,\ldots,\hat{\mathbf{v}}_k\}$.
We define the projection onto this subspace as
\begin{equation}
\mathbf{P}
= 
\mathbf{V}(\mathbf{V}^{\top}\mathbf{V})^{-1}\mathbf{V}^{\top}.
\end{equation}

During training, we cache the base-model hidden states $\mathbf{h}_0$ at the selected layer and compare them with the adapted model hidden states $\mathbf{h}_\theta$ at the same position. For each training example $i$, we adapt the regularization weight according to the adapted model's probability on the correct answer:
\begin{equation}
\lambda_i = p_\theta\!\left(y_i^{\mathrm{gold}} \mid x_i\right),
\end{equation}
The training loss is
\begin{equation}
\mathcal{L}_i
=
(1-\lambda_i)\mathcal{L}_{\mathrm{CE},i}
+
\lambda_i
\left\|
(\mathbf{I}-\mathbf{P})
(\mathbf{h}_i-\mathbf{h}_{0,i})
\right\|_2^2.
\end{equation}
The regularization term penalizes changes to model activations in directions orthogonal to the ambiguity subspace, while allowing the update to freely change activations within the span of the ambiguity directions. Examples for which the model assigns high probability to the correct answer are strongly guided toward updates within the ambiguity subspace, whereas examples assigned low probability receive greater weight on the task objective.

We target the attention \texttt{q\_proj} and
\texttt{v\_proj} modules\footnote{\texttt{Phi-4-mini-instruct} uses a combined \texttt{qkv\_proj} module rather than separate \texttt{q\_proj} and \texttt{v\_proj} modules.} and train for three epochs. We sweep over LoRA ranks $r \in \{4,8,16\}$ with matching scaling factors,
and learning rates 
$\{10^{-4},10^{-5}\}$ resulting in 6 configurations per model. We provide additional details on training data and hyperparameter selection in~App.~\ref{app:evaluation}.

\paragraph{Baselines.} We compare DART against the base (unmodified instruction-tuned) model.

To measure how much of DART's performance is attributable to fine-tuning itself, we compare to LoRA with no regularizer. This is equivalent to DART with $\lambda=0$. To measure how much of DART's performance is attributable to the use of ambiguity vectors, we also compare against DART applied to a random subspace of the same rank.

\begin{figure*}[t]
    \centering
    \includegraphics[width=\linewidth]{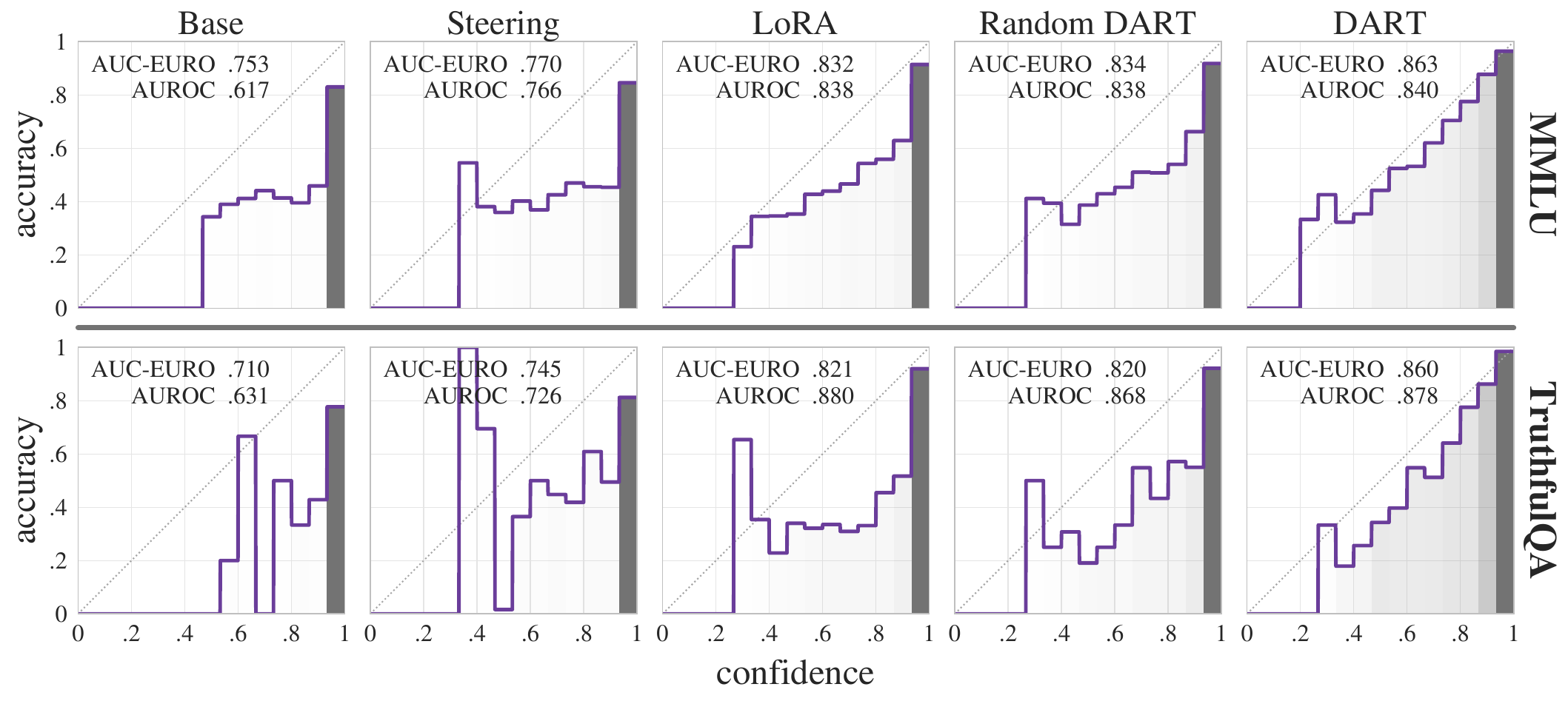} 
    \caption{\textbf{Reliability diagrams for} \texttt{Llama-3.3-70B-Instruct} under the logits protocol. 
    Results are shown for the base model, ambiguity steering, LoRA, and DART on MMLU (top) and TruthfulQA (bottom). DART substantially improves calibration while maintaining accuracy. We use the best hyperparameter configuration for each method, as described in~App.~\ref{app:best_hyperparam} and as selected in Fig.~\ref{fig:pareto_logits_70B}.
    }
    \label{fig:rel_grid_70B}
\end{figure*}

\section{Results}
\label{sec: results}
\paragraph{Ambiguity regularization improves calibration.}
We first study whether DART improves confidence estimation while preserving task performance. Figure~\ref{fig:rel_grid_70B} shows reliability diagrams for the selected configurations of each method, with histograms indicating the distribution of confidence estimates relative to accuracy. The primary failure mode of base LMs under logits-based confidence elicitation is overconfidence, which is only marginally improved by steering (\cref{fig:pareto_logits_70B}). Standard LoRA reduces miscalibration, while DART further reduces it and better aligns confidence with empirical accuracy across a broader range of confidence values, with little change in task performance. These trends hold across models of different families and sizes, with DART consistently achieving better calibration over standard LoRA on multiple-choice QA tasks (App.~\ref{app:detailed_results}). 

\paragraph{The role of ambiguity regularization.}
While standard LoRA already improves calibration, DART provides additional gains, suggesting that the identified ambiguity subspace contributes to calibration beyond standard fine-tuning. To test this hypothesis, we replace the ambiguity subspace with a random subspace of the same rank as that used in DART, while keeping the fine-tuning procedure unchanged. As shown in Figures~\ref{fig:rel_grid_70B}~and~\ref{fig:random_dart}, regularizing toward a random subspace also improves confidence estimation relative to the base model, yet DART with the ambiguity directions achieves even lower calibration error and higher AUC-EURO on both datasets.

\paragraph{Beyond multiple-choice QA and logit-based confidence.} 
To assess whether these findings generalize to open-ended generation tasks, we additionally evaluate our methods on AmbigQA \citep{min-etal-2020-ambigqa}. As shown in~\cref{tab:ambigqa-logits-70B}, we observe similar patterns as in multiple-choice settings: steering provides modest improvements, while fine-tuning with DART substantially improves calibration with little change in task performance. 

We further investigate whether verbalized confidence estimates show similar trends. Under this protocol, confidence estimates are far less well calibrated than under logits-based estimation. Improvements are also generally smaller than for logits-based confidence, consistent with prior work showing that the relative performance of confidence elicitation protocols depends greatly on the task setting and model~\citep[cf.][]{tian-etal-2023-just,shrivastava2023llamasknowgptsdont}. Detailed results under verbalized confidence for multiple-choice and open-ended generation are shown in~App.~\ref{app:detailed_results}. 

\begin{figure*}
    \centering
    \includegraphics[width=\textwidth]{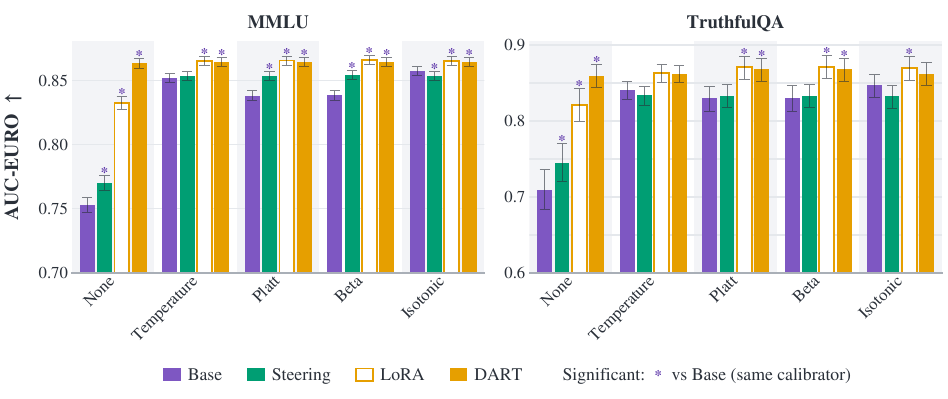}
    \caption{\textbf{Post-hoc calibration for} \texttt{Llama-3.3-70B-Instruct} under the \textbf{logits} protocol.
    AUC-EURO is shown for the base model, steering, LoRA, and DART, as trained (None) and after temperature scaling, Platt scaling, beta calibration, and isotonic regression, on MMLU (left) and TruthfulQA (right). Error bars show 95\% bootstrap intervals. * marks a significant difference from the base model under the same calibrator.}
    \label{fig:post-hoc-70B-main}
\end{figure*}

\paragraph{Post-hoc calibration complements these methods.}
Finally, we evaluate temperature scaling, Platt scaling, beta calibration, and isotonic regression. 
These methods simply adjust existing confidence estimates without modifying the underlying model, so task performance remains unchanged.  
~\Cref{fig:post-hoc-70B-main} shows that these methods substantially reduce calibration error.
Note that post-hoc calibration is orthogonal to representational methods, and that both can be combined. Thus, we also try combining post-hoc calibration with the methods we have evaluated, including steering, LoRA, and DART.
Further improvements beyond post-hoc methods can often be achieved by combining post-hoc calibration methods with LoRA or DART. Further details are provided in  Appendix~\ref{app:cross-model-results-ts}.

\begin{table}[]
    \small
    \centering
    \setlength{\tabcolsep}{3pt}
    \begin{tabular}{lcccc}
    \toprule
    Method & smECE\,$\downarrow$ & AUC-EURO\,$\uparrow$ & EM\,$\uparrow$ & F1\,$\uparrow$ \\
    \midrule
    Baseline    & 0.543 & 0.540 & 0.426 & 0.554 \\
    \midrule
    Steering    & 0.458 & 0.599 & 0.370 & 0.504 \\
    LoRA        & 0.127 & 0.777 & \textbf{0.519} & \textbf{0.628} \\
    Random DART & 0.136 & 0.774 & 0.503 & 0.614 \\
    DART        & \textbf{0.090} & \textbf{0.778} & 0.485 & 0.596 \\
    \bottomrule
    \end{tabular}
    \caption{Calibration of \texttt{Llama-3.3-70B-Instruct} on AmbigQA.}
    \label{tab:ambigqa-logits-70B}
\end{table}

\section{Conclusion}
\label{sec: conclusion}
We have proposed a protocol for locating representations of ambiguity in language models, and using these representations to guide fine-tuning.
By encouraging model updates to lie in the span of the ambiguity representations, our proposed fine-tuning approach, DART, yields consistent improvements in confidence estimation across multiple models while incurring little to no loss in accuracy on multiple-choice QA and open-ended generation QA datasets. 

\section*{Limitations}

Ambiguity has many forms, and our operationalization of ambiguity only captures one component of it (multiple interpretations of a given input). We have not considered ambiguity as operationalized by the vagueness of an open-domain question. Furthermore, positive results with DART do not necessarily indicate that we have recovered representations of ambiguity \emph{per se}. Any interpretability conclusions that relate to whether LMs cleanly represent ambiguity should be taken as preliminary.

We have not explicitly contrasted representations of ambiguity, uncertainty, and difficulty. Each of these could mediate confidence estimations in language models in different ways; combining them could potentially yield greater improvements in confidence estimation. We have shown that ambiguity representations can modulate this factor, but we have not shown that ambiguity is necessarily better for this than other concept representations. As isolating representations of uncertainty and difficulty is more difficult and potentially more dependent on the specific model being evaluated,
we did not investigate these phenomena. 

Although we show that confidence estimation can be improved at inference-time or by DART with ambiguity steering vectors applied to LoRA, these protocols are not a replacement for careful, human-in-the-loop risk assessment. Users should always exercise caution when trusting language model output.

\section*{Acknowledgments}
We thank Sheridan Feucht, Michael Li, and members of the
BAAIGL and MIILK groups at Boston University for helpful discussions
and constructive feedback.

\bibliography{custom, anthology-1, anthology-2}

\clearpage
\appendix
\section{Ambiguity Dataset Construction }
\label{app:Ambient}
Figure~\ref{fig:ambient-examples} shows the procedure we follow to create our dataset from \textsc{AmbiEnt}. Every ambiguous sentence is paired with one or more disambiguations, each resolving the ambiguity differently. For each disambiguation, we construct a contrast pair: the original ambiguous sentence, and a clear sentence minimally edited to make that reading explicit. This keeps the two sentences identical except for the information needed to resolve the ambiguity.

\begin{figure*}[p]
\centering
\begin{minipage}{\textwidth}

\ex.\label{ex:pragmatic}
\textbf{Original \textsc{AmbiEnt} sentence:} I'm afraid \underline{the cat was hit by a car}. \hfill\textit{Pragmatic}
\a.
\begin{tabular}[t]{@{}ll@{}}
\multicolumn{2}{@{}l}{\textbf{Contrastive pair 1}}\\[0.2em]
\textbf{Ambiguous:}
    & I'm afraid the cat was hit by a car. \\
\textbf{Disambiguated:}
    & I'm \underline{\textbf{\textit{worried}}} the cat was hit by a car.
\end{tabular}
\hspace{1.5cm}
\b.
\begin{tabular}[t]{@{}ll@{}}
\multicolumn{2}{@{}l}{\textbf{Contrastive pair 2}}\\[0.2em]
\textbf{Ambiguous:}
    & I'm afraid the cat was hit by a car. \\
\textbf{Disambiguated:}
    & I'm \underline{\textbf{\textit{sorry to share that}}} the cat was hit by a car.
\end{tabular}

\ex.\label{ex:lexical}
\textbf{Original \textsc{AmbiEnt} sentence:} John and Anna are \underline{married}. \hfill\textit{Lexical}
\a.
\begin{tabular}[t]{@{}ll@{}}
\multicolumn{2}{@{}l}{\textbf{Contrastive pair 1}}\\[0.2em]
\textbf{Ambiguous:}
    & John and Anna are married. \\
\textbf{Disambiguated:}
    & John and Anna are \underline{\textbf{\textit{both}}} married.
\end{tabular}
\hspace{1.5cm}
\b.
\begin{tabular}[t]{@{}ll@{}}
\multicolumn{2}{@{}l}{\textbf{Contrastive pair 2}}\\[0.2em]
\textbf{Ambiguous:}
    & John and Anna are married. \\
\textbf{Disambiguated:}
    & John and Anna are married \underline{\textbf{\textit{to each other}}}.
\end{tabular}

\ex.\label{ex:syntactic}
\textbf{Original \textsc{AmbiEnt} sentence:} There will be \underline{more interesting seminars}\ldots \hfill\textit{Syntactic}
\a.
\begin{tabular}[t]{@{}ll@{}}
\multicolumn{2}{@{}l}{\textbf{Contrastive pair 1}}\\[0.2em]
\textbf{Ambiguous:}
    & There will be more interesting seminars\ldots \\
\textbf{Disambiguated:}
    & There will be \underline{\textbf{\textit{more}}} seminars\ldots that are interesting.
\end{tabular}
\hspace{1.5cm}
\b.
\begin{tabular}[t]{@{}ll@{}}
\multicolumn{2}{@{}l}{\textbf{Contrastive pair 2}}\\[0.2em]
\textbf{Ambiguous:}
    & There will be more interesting seminars\ldots \\
\textbf{Disambiguated:}
    & There will be seminars\ldots that are \underline{\textbf{\textit{more interesting}}}.
\end{tabular}

\ex.\label{ex:scopal}
\textbf{Original \textsc{AmbiEnt} sentence:} The novel has \underline{not been banned in many schools}. \hfill\textit{Scopal}
\a.
\begin{tabular}[t]{@{}ll@{}}
\multicolumn{2}{@{}l}{\textbf{Contrastive pair 1}}\\[0.2em]
\textbf{Ambiguous:}
    & The novel has not been banned in many schools. \\
\textbf{Disambiguated:}
    & There are many schools where the novel has \underline{\textbf{\textit{not been banned}}}.
\end{tabular}
\hspace{1.5cm}
\b.
\begin{tabular}[t]{@{}ll@{}}
\multicolumn{2}{@{}l}{\textbf{Contrastive pair 2}}\\[0.2em]
\textbf{Ambiguous:}
    & The novel has not been banned in many schools. \\
\textbf{Disambiguated:}
    & \underline{\textbf{\textit{It is not the case}}} that the novel has been banned in many schools.
\end{tabular}

\end{minipage}
\caption{Examples of the four ambiguity types in our dataset: \textit{pragmatic, lexical, syntactic,} and \textit{scopal}.}
\label{fig:ambient-examples}
\end{figure*}

Table~\ref{tab:category-train-val-counts} reports the number of paired
ambiguity contrasts in each category after converting \textsc{AmbiEnt} examples into
matched ambiguous--clear sentence pairs. The four categories we use are \textit{pragmatic, lexical, scopal} and \textit{syntactic} ambiguity. We aim for at least 50 examples in the training set after splitting; the remaining categories from \textsc{AmbiEnt} are filtered out, as they do not meet this criterion.

\begin{table}[ht]
\centering
\small
\begin{tabular}{lrr}
\toprule
\textbf{Category} & \textbf{Train} & \textbf{Validation} \\
\midrule
Pragmatic     & 497  & 124 \\
Lexical       & 244  & 61  \\
Scopal        & 90   & 22  \\
Syntactic     & 76   & 19  \\
\midrule
\textbf{Total} & \textbf{907} & \textbf{226} \\
\bottomrule
\end{tabular}
\caption{Number of examples in the training and validation splits for each ambiguity category.}
\label{tab:category-train-val-counts}
\end{table}

\FloatBarrier
\section{Prompt and Tokenization Details}
\label{app:prompts}

All inputs are formatted as a single user message using the corresponding
model's chat template with a generation prompt appended. After chat formatting, we do not add additional tokenizer special tokens. This same formatting convention is
used for ambiguity direction construction, layer selection, inference time steering, LoRA activation caching, LoRA training, and downstream evaluation.

\paragraph{Ambiguity judgment prompt.}
To construct ambiguity directions and select the intervention layer, we use
the prompt template in Figure~\ref{fig:amb-judgment-prompt}, replacing
\texttt{\{sentence\}} with the input sentence.

\begin{figure}[h]
\centering
\begin{promptbox}
\small
Decide whether the sentence is ambiguous.\\
A sentence is ambiguous if, without extra context, it can be interpreted in
more than one distinct plausible way.\\
Answer Yes if the sentence is ambiguous, and No otherwise.\\
Reply with exactly Yes or No.\\

Examples:\\
Sentence: We reached the bank.\\
Answer: Yes\\

Sentence: The chicken is ready to eat.\\
Answer: Yes\\

Sentence: Every student read the same book.\\
Answer: No\\

Sentence: I saw the man with the telescope.\\
Answer: Yes\\

Sentence: The chicken is ready to be eaten.\\
Answer: No\\

Sentence: We reached the river bank.\\
Answer: No\\

Sentence: Every student read a book.\\
Answer: Yes\\

Sentence: I saw the man through the telescope.\\
Answer: No\\

Sentence: \texttt{\{sentence\}}\\
Answer:
\end{promptbox}
\caption{Prompt template used for ambiguity judgment, ambiguity direction
construction, and intervention layer selection.}
\label{fig:amb-judgment-prompt}
\end{figure}

\paragraph{Logits prompt.}
For logits based confidence, we use the prompt template shown in
\cref{fig:logits-prompt} for each MMLU and TruthfulQA example, replacing the
placeholders with the corresponding question and answer choices.

\begin{figure}[h]
\centering
\begin{promptbox}
\small
Answer the following multiple-choice question.\\

Question: \texttt{\{question\}}\\
A. \texttt{\{choice A\}}\\
B. \texttt{\{choice B\}}\\
C. \texttt{\{choice C\}}\\
D. \texttt{\{choice D\}}\\

Give your final answer only. Do not explain, show work, or include reasoning.\\
Respond with exactly one letter (A, B, C, or D) and nothing else.\\
Answer:
\end{promptbox}
\caption{Prompt template used for logits-based confidence on MMLU and
TruthfulQA examples.}
\label{fig:logits-prompt}
\end{figure}

\paragraph{Verbalized confidence prompt.}
For verbalized confidence, we use the same question format, but ask the
model to report both an answer and a confidence value, as shown in
\cref{fig:verbalized-confidence-prompt}.

\begin{figure}[h]
\centering
\begin{promptbox}
\small
Answer the following multiple-choice question.\\

Question: \texttt{\{question\}}\\
A. \texttt{\{choice A\}}\\
B. \texttt{\{choice B\}}\\
C. \texttt{\{choice C\}}\\
D. \texttt{\{choice D\}}\\

Report Confidence as an integer from 0 to 100 for how likely your answer is to be correct.\\
Give your final answer only. Do not explain, show work, or include reasoning.\\
Respond with exactly two lines and nothing else:\\
Answer: \texttt{<A/B/C/D>}\\
Confidence: \texttt{<0-100>}\\

Answer:
\end{promptbox}
\caption{Prompt template used for verbalized confidence on MMLU and
TruthfulQA examples.}
\label{fig:verbalized-confidence-prompt}
\end{figure}

\FloatBarrier
\section{Validation and Evaluation Details}
\label{app:evaluation}
\subsection{MMLU}
We evaluate on all subjects of MMLU \citep{hendryckstest2021}. LoRA and DART adapters are trained on a fixed subset of 5,000 examples from MMLU \texttt{auxiliary\_train} split. For each model method and confidence elicitation protocol, we select the configuration that maximizes AUC-EURO on the validation split and report results on the test split. Post-hoc calibrators are fitted on the validation split and evaluated on the test split.

\subsection{TruthfulQA}
We additionally evaluate on the MC1 subset of TruthfulQA \citep{lin2021truthfulqa}. To standardize the number of answer choices, we only keep examples with at least four options, resulting in 684 of the original 817 questions. For the questions with exactly four options, we keep all choices. For the remaining questions, we keep the correct answer and randomly sample three incorrect choices. Lastly, we shuffle the answer choices.

Since TruthfulQA provides a single split, we use repeated 10-fold cross fitting with 20 repetitions. For this dataset, we used the adapters trained on MMLU, without additional fine-tuning. For each fold, we select the configuration that maximizes AUC-EURO on the other nine and evaluate it on the held-out fold. We repeat the procedure independently for each model, method and confidence elicitation protocol, and aggregate held-out predictions across all repetitions to compute the reported metrics. We follow a similar procedure for post-hoc calibration, where we fit the calibrators on the other nine folds using the selected configuration and apply them to the held-out fold.

\subsection{AmbigQA}
We evaluate open-ended generation on AmbigQA \citep{min-etal-2020-ambigqa}. Since the test split is not publicly available, we reserve 5,036 questions form the training split for evaluation and use the remaining 5,000 for fine-tuning, using stratified sampling by ambiguity. For fine-tuning, questions
with multiple interpretations contribute one target answer per interpretation, each weighted by the inverse of the number of interpretations. We select hyperparameters to maximize AUC-EURO on the validation split. We use exact match (EM) against the annotated answers to determine correctness for confidence estimation metrics and report token-level F1 alongside to measure task performance.

To mirror the logits-based confidence protocol used for multiple-choice tasks, we derive confidence from the likelihood assigned to the generated answer. We greedily decode and define confidence as the likelihood normalized by length
\begin{equation}
\exp\left(
\frac{1}{T}\sum_{t=1}^{T}
\log p_\theta(y_t \mid x,y_{<t})
\right),
\end{equation}
where \(T\) is the number of generated tokens. We refer to this method as Likelihood.

\subsection{Statistical analysis}
\label{app:statistical_analysis}
We compare each method with the base model under the same confidence elicitation protocol using two-sided paired permutation tests over questions. The tests use $5,000$ permutations for MMLU and AmbigQA and $1,000$ for TruthfulQA, where cross-fitted configuration selection is repeated within each permutation. We apply Holm–Bonferroni correction \citep{holm1979simple} at $\alpha=0.05$ separately for each model, method, and confidence protocol. Each correction covers nine tests evaluating smECE, AUC-EURO, and AUROC on the three datasets.

We also report $95\%$ percentile bootstrap confidence intervals based on $5,000$ resamples of the evaluation questions. The selected configurations remain fixed during resampling, and all $20$ cross-fitted predictions for each TruthfulQA question are kept together.

\FloatBarrier
\section{Selected Hyperparameters}
\label{app:best_hyperparam}

\Cref{tab:best_configurations} reports the hyperparameter configurations selected for \texttt{Llama-3.3-70B-Instruct}. For MMLU and AmbigQA, we select configurations that maximize AUC-EURO on their respective validation splits. For TruthfulQA, configurations are selected separately for each held-out fold following the cross-fitting procedure described in~App.~\ref{app:evaluation}. We report the most frequently selected configuration across folds.

\begin{table}[H]
    \centering
    \setlength{\tabcolsep}{3pt}
    \begin{tabular*}{\columnwidth}{@{\extracolsep{\fill}}llccc@{}}
    \toprule
    \textbf{Protocol} & \multicolumn{2}{c}{\textbf{Steering}}
        & \textbf{LoRA} & \textbf{DART} \\
    \cmidrule(lr){2-3}
        & \textbf{Vector} & \textbf{Strength} & $r$ & $r$ \\
    \midrule
    \multicolumn{5}{@{}l@{}}{\textit{MMLU}} \\
    Logit      & Scopal    & 200\% & 8  & 8  \\
    Verbalized & Pragmatic & 50\%  & 8  & 16 \\
    \midrule
    \multicolumn{5}{@{}l@{}}{\textit{TruthfulQA}} \\
    Logit      & Lexical   & 200\% & 8  & 16 \\
    Verbalized & Pragmatic & 100\% & 8  & 4  \\
    \midrule
    \multicolumn{5}{@{}l@{}}{\textit{AmbigQA}} \\
    Likelihood & Syntactic & 200\% & 16 & 8  \\
    Verbalized & Pragmatic & 100\% & 4  & 4  \\
    \bottomrule
    \end{tabular*}
    \caption{Selected hyperparameter configurations for \texttt{Llama-3.3-70B-Instruct}. The LoRA and DART columns report rank $r$; for both methods, all reported configurations have LoRA scaling parameter $\alpha=r$ and learning rate $10^{-5}$. Steering strength is expressed as a percentage.}
    \label{tab:best_configurations}
\end{table}

\FloatBarrier
\section{Additional Selectivity Results}
\label{app:model-offset-selectivity}
In this section, we present additional selectivity results.

\subsection{Selectivity Across Models}
\begin{figure*}[!b]
    \centering
    \includegraphics[width=\textwidth]{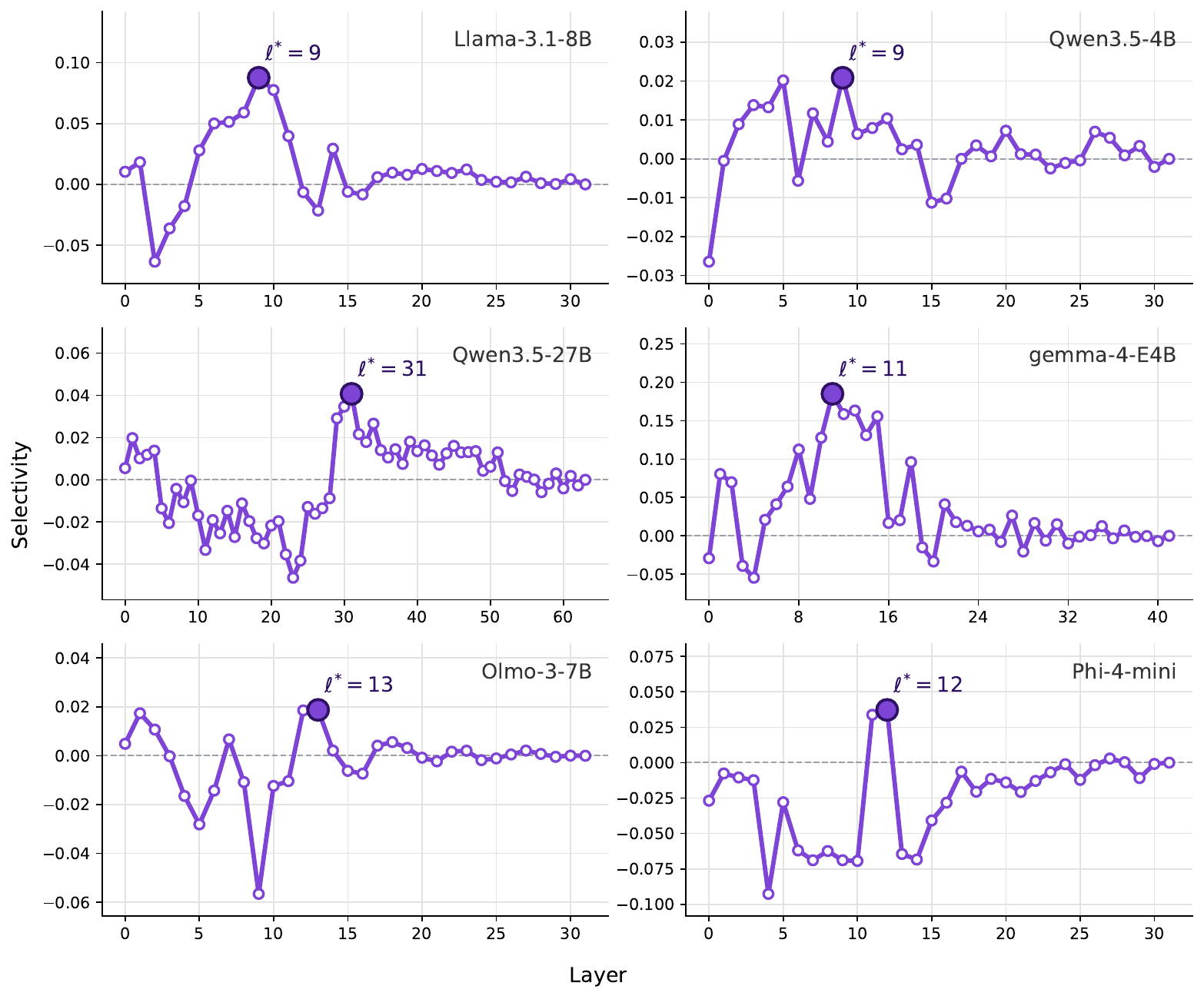}
    \caption{Selectivity of candidate ambiguity directions across layers. Weighted mean selectivity is averages over relative strengths and core ambiguity categories for each model.}
    \label{fig:selectivity-across-models-single-offset}
\end{figure*}

In \cref{fig:selectivity-across-models-single-offset}, we show selectivity results for the remaining evaluated models. For each model, we select the layer with highest selectivity as the one to perform interventions. Across all models, this peak falls in the early-to-mid layers, suggesting that ambiguity is most cleanly represented relatively early in the network.

\subsection{Additional selectivity offsets}
\begin{figure*}[!t]
    \centering
    \includegraphics[width=\textwidth]{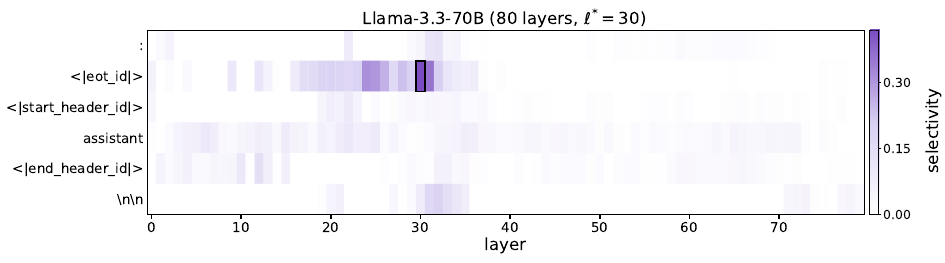}
    \caption{Selectivity across offsets for \texttt{Llama-3.3-70B-Instruct}.}
    \label{fig:selectivity-offsets-70B}
\end{figure*}

\begin{figure*}[!t]
    \centering
    \includegraphics[width=\textwidth]{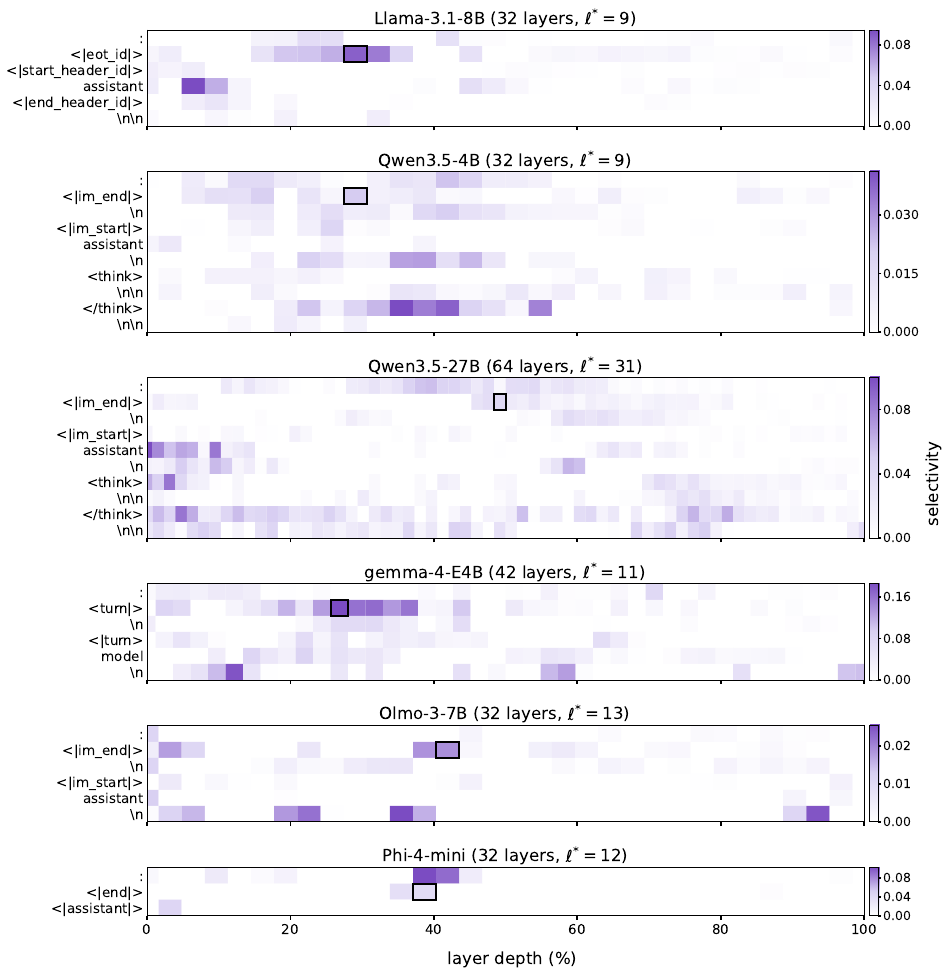}
    \caption{Selectivity across offsets for the remaining evaluated models.}
    \label{fig:selectivity-offsets-rest}
\end{figure*}

We define candidate positions after applying the model's chat template. We
identify the final colon in the last \texttt{Answer:} cue and consider
nonnegative offsets from that token. Offset 0 refers to the colon itself, while
positive offsets refer to subsequent prompt tokens, including any
model specific control tokens inserted by the chat template before generation
begins. For each candidate position, we extract hidden states from the decoder
layer outputs.
In \cref{fig:selectivity-offsets-70B,fig:selectivity-offsets-rest} we show selectivity results for different token positions across all models. While the position of peak selectivity varies by model and token, we observe that middle layers consistently show high selectivity across models. We fix one global offset (offset 1) as it reliably falls within this middle-layer band for all models.

\clearpage
\twocolumn

\section{Additional Qualitative Evaluation}
\label{app:additional-qual-eval}
In \cref{tab:dalai-lama-qualitative-full,tab:spilled-beans-qualitative-full} we show full qualitative evaluation examples showing the effects of the ambiguity interventions on different datasets and prompt formats. The examples support the trend observed in the main text: adding the ambiguity vector makes the model more sensitive to alternative readings of an input, while subtracting it leads the model toward a single, confident interpretation. This pattern holds both for the multiple-choice format of MMLU and for an open-ended interpretation question drawn from \textsc{AmbiEnt}.

\begin{table*}[!b]
\footnotesize
\centering
\setlength{\fboxsep}{5pt}

\begin{minipage}{0.98\textwidth}

\noindent\colorbox{promptcolor}{%
\begin{minipage}{\dimexpr\linewidth-2\fboxsep\relax}
\raggedright
\textbf{Prompt:}

\vspace{0.25em}
What does the following sentence mean? 

\emph{Maria spilled the beans.}
\end{minipage}}

\genbox{basecolor}{No intervention}{

The sentence ``Maria spilled the beans'' is an \textbf{idiom} that means
Maria revealed a secret or disclosed confidential information
that she was supposed to keep private.

\vspace{0.2em}
Here is a breakdown of the meaning:
\begin{itemize}
    \item \textbf{Literal meaning:} It refers to the act of accidentally
    dropping beans from a pod, which would be messy and wasteful.
    \item \textbf{Figurative meaning:} In this context, it implies that Maria
    accidentally or sometimes intentionally let someone in on a surprise, a
    plan, or a piece of gossip that others did not know about.
\end{itemize}

Example usage:
``We were all planning a big surprise party for Maria's birthday, but she
spilled the beans to her roommate, so now everyone knows.''
}

\genbox{poscolor}{Adding ambiguity vector}{

The sentence ``Maria spilled the beans'' is an \textbf{idiom} that means
\textbf{Maria revealed a secret or shared confidential information.}

\vspace{0.2em}
While the literal action of spilling physical beans, like from a bag, is what
the phrase describes, \textbf{it can also be used metaphorically} to indicate that
someone has disclosed something they were supposed to keep private.

\vspace{0.2em}
Meaning in this context:
When someone says ``Maria spilled the beans,'' it usually means
Maria revealed a secret.
}

\genbox{negcolor}{Subtracting ambiguity vector}{

The sentence ``Maria spilled the beans'' \textbf{is a metaphorical expression.}

\vspace{0.2em}
In this context, the phrase means that Maria revealed a secret or
disclosed confidential information.

\vspace{0.2em}
The idiom ``spilling the beans'' refers to the act of revealing hidden
truths or exposing secrets. It suggests that Maria was
unhiding information or making a disclosure, much like
someone in a laboratory or a mystery novel.

\vspace{0.2em}
Therefore, the sentence implies that Maria was the one who disclosed
the truth or uncovered the secret.
}

\end{minipage}

\caption{Full qualitative evaluation of the ambiguity vector on an \textsc{AmbiEnt} example exhibiting \textit{idiomatic} ambiguity. Adding the vector preserves sensitivity to multiple
readings of \emph{Maria spilled the beans}, while subtracting it leads the
model toward a single interpretation.}
\label{tab:spilled-beans-qualitative-full}
\end{table*}

\begin{table*}[p]
\footnotesize
\centering
\setlength{\fboxsep}{5pt}

\begin{minipage}{0.98\textwidth}

\noindent\colorbox{promptcolor}{%
\begin{minipage}{\dimexpr\linewidth-2\fboxsep\relax}
\raggedright
\textbf{Prompt:}

\vspace{0.25em}
Answer the following multiple-choice question.

\vspace{0.3em}
Question: When was the \emph{current Dalai Lama} born?

\vspace{0.2em}
A. 1935 \quad
B. 2000 \quad
C. 1965 \quad
D. 1975

\vspace{0.3em}
Briefly explain your interpretation and reasoning. If the question seems
ambiguous or underspecified, mention why. Then give your final answer and
confidence in this format:

\vspace{0.2em}
\texttt{Answer: <A/B/C/D>} 

\texttt{Confidence: <0--5>}
\end{minipage}}

\genbox{basecolor}{No intervention}{

The question asks for the birth year of the ``current Dalai Lama.'' This
refers to the 14th Dalai Lama, Tenzin Gyatso. He was born on June 6, 1935,
in Lhasa, Tibet.

\vspace{0.2em}
Looking at the options:
\begin{itemize}[noitemsep]
    \item A. 1935: Matches the historical record.
    \item B. 2000: Incorrect; this is in the future relative to his birth.
    \item C. 1965: Incorrect; this would make him 30 years old at birth,
    which is not the case.
    \item D. 1975: Incorrect; this is also in the future relative to his birth.
\end{itemize}

There is \textbf{no ambiguity} regarding the birth year of the 14th Dalai
Lama in historical records.

\vspace{0.2em}
\upshape Answer: A \\
Confidence: 5
}

\genbox{poscolor}{Adding ambiguity vector}{

The question asks for the birth year of the ``current Dalai Lama.'' This
phrasing can be \textbf{slightly ambiguous because the title ``Dalai Lama''
has been held by different individuals at different times.}

\vspace{0.2em}
\begin{enumerate}[noitemsep]
    \item The 14th Dalai Lama: Tenzin Gyatso is the most widely
    recognized figure holding this title currently. He was born on June 6,
    1935. He has been the spiritual leader of Tibetan Buddhism and
    the political leader of the Tibetan exile community since 1950.

    \item The 15th Dalai Lama: In 2011, the 14th Dalai Lama selected
    a new successor, the 15th Dalai Lama, named Lobsang Yangzhen. He was born
    in \textbf{1965}. However, he is not yet the reigning sovereign or the
    universally recognized ``current'' Dalai Lama in the same capacity as the
    14th, as the 14th Dalai Lama remains the primary figurehead.
\end{enumerate}

Given that multiple-choice questions on general knowledge typically refer to
the longest-serving or most prominent holder of a title unless specified
otherwise, the question almost certainly refers to the 14th Dalai Lama,
Tenzin Gyatso.

\vspace{0.2em}
\begin{itemize}[noitemsep]
    \item Option A (1935) matches the birth year of the 14th Dalai Lama.
    \item Option B (2000) is incorrect; no Dalai Lama was born then.
    \item Option C (1965) matches the birth year of the 15th Dalai Lama, but
    he is not the primary ``current'' holder in standard historical context.
    \item Option D (1975) is incorrect.
\end{itemize}

Therefore, the intended answer is 1935.

\vspace{0.2em}
\upshape Answer: A \\
Confidence: 4
}

\genbox{negcolor}{Subtracting ambiguity vector}{

The question asks for the birth year of the ``current Dalai Lama.'' \textbf{This
refers to the 14th Dalai Lama, Tenzin Gyatso. He was born on June 6, 1935,
in Lhasa, Tibet.}

\vspace{0.2em}
Looking at the options:
\begin{itemize}[noitemsep]
    \item A. 1935: Matches the historical record.
    \item B. 2000: Incorrect; this is in the future relative to his birth.
    \item C. 1965: Incorrect; this would make him 30 years old at birth,
    which is impossible.
    \item D. 1975: Incorrect; this is also in the future relative to his birth.
\end{itemize}

The question is \textbf{not ambiguous} regarding the identity of the person
\textbf{(the 14th Dalai Lama is the recognized current one)}, and the birth year is a
well-documented historical fact.

\vspace{0.2em}
\upshape Answer: A \\
Confidence: 5
}

\end{minipage}

\caption{Full qualitative evaluation of the ambiguity vector on an MMLU example. Steering positively, makes the model sensitive to the \emph{current} Dalai Lama, while subtracting it leads to a single interpretation.}
\label{tab:dalai-lama-qualitative-full}

\end{table*}

\FloatBarrier
\section{Results Across Models and Confidence Elicitation Protocols}
\label{app:detailed_results}
\begin{figure*}[]
    \centering
    \includegraphics[width=\textwidth]{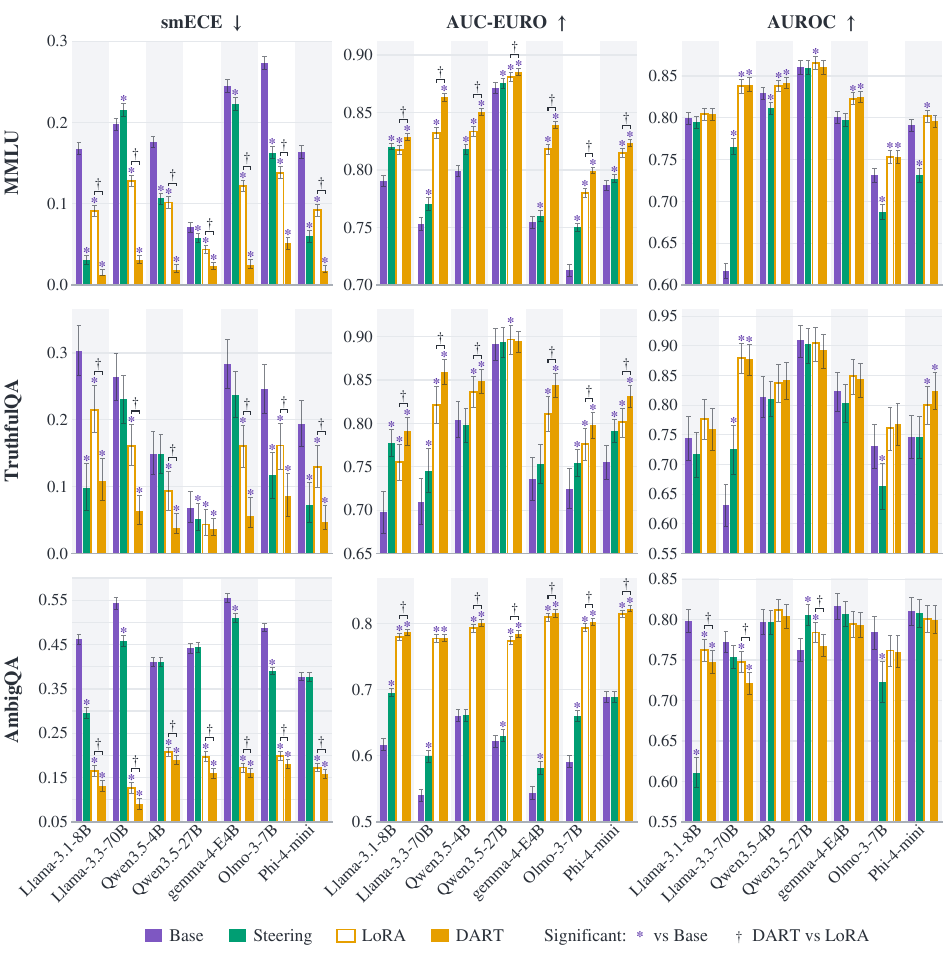}
    \caption{\textbf{Calibration results for all models} under the \textbf{logits} protocol.
    smECE, AUC-EURO, and AUROC are shown for the base model, steering, LoRA, and DART on MMLU (top), TruthfulQA (middle), and AmbigQA (bottom). Error bars show 95\% bootstrap intervals. $*$ marks a statistically significant difference from the base model,
    and $\dagger$ marks a statistically significant difference between DART and LoRA. Selected configurations are listed in~App.~\ref{app:best_hyperparam}.}
    \label{fig:calibration_logit}
\end{figure*}

\begin{figure*}[]
    \centering
    \includegraphics[width=\textwidth]{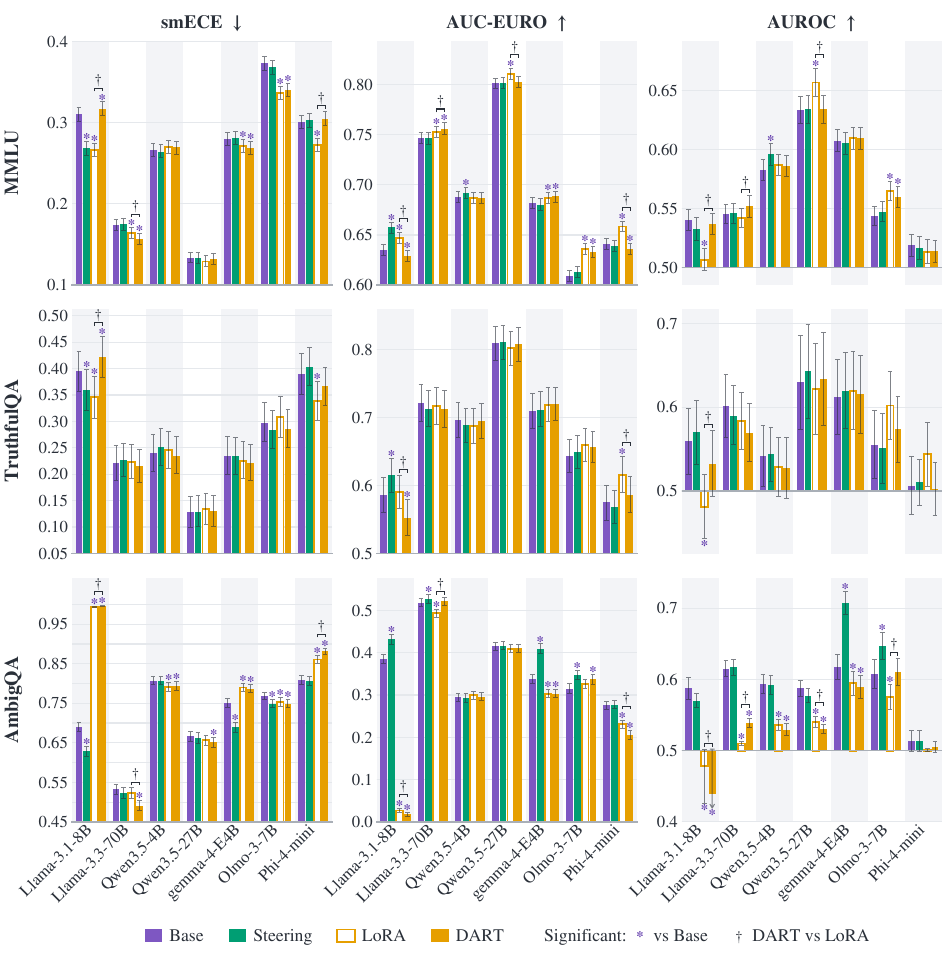}
    \caption{\textbf{Calibration results for all models} under the \textbf{verbalized} protocol.
    smECE, AUC-EURO, and AUROC are shown for the base model, steering, LoRA, and DART on MMLU (top), TruthfulQA (middle), and AmbigQA (bottom). Error bars show 95\% bootstrap intervals. $*$ marks a statistically significant difference from the base model,
    and $\dagger$ marks a statistically significant difference between DART and LoRA. Selected configurations are listed in~App.~\ref{app:best_hyperparam}.}
    \label{fig:calibration_verbalized}
\end{figure*}

In addition to the logit-based confidence results, we evaluate our methods under the verbalized confidence elicitation protocol described in~\cref{sec:confidence}. 
In~\cref{fig:calibration_logit,fig:calibration_verbalized}, we show results across models, tasks and confidence elicitation protocols, which follow broadly consistent trends with the main results.
Under the verbalized confidence protocol, parsing failures can affect task performance. As described in~\cref{sec:confidence}, generations that do not contain both a valid answer and a valid confidence estimate are treated as incorrect predictions with maximum confidence. This issue is particularly pronounced for \texttt{Llama-3.1-8B-Instruct} after both LoRA and DART on AmbigQA.

\begin{figure*}[]
    \centering
    \includegraphics[width=\textwidth]{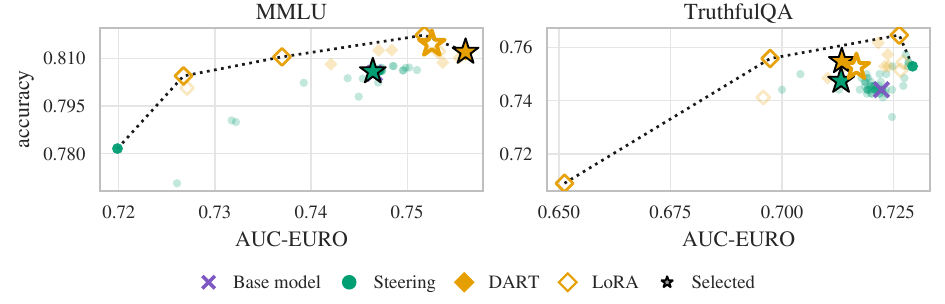}
    \caption{\textbf{Trade-off between AUC-EURO and accuracy under verbalized confidence elicitation on MMLU (left) and TruthfulQA (right)}. Results shown for \texttt{Llama-3.3-70B-Instruct}. Each point represents one hyperparameter configuration. Stars mark the selected configuration for each method. Higher values are better on both axes. The dotted line shows the Pareto frontier.}
    \label{fig:pareto_verbalized_euro_70B}
\end{figure*}

\begin{figure*}[]
    \centering
    \includegraphics[width=\textwidth]{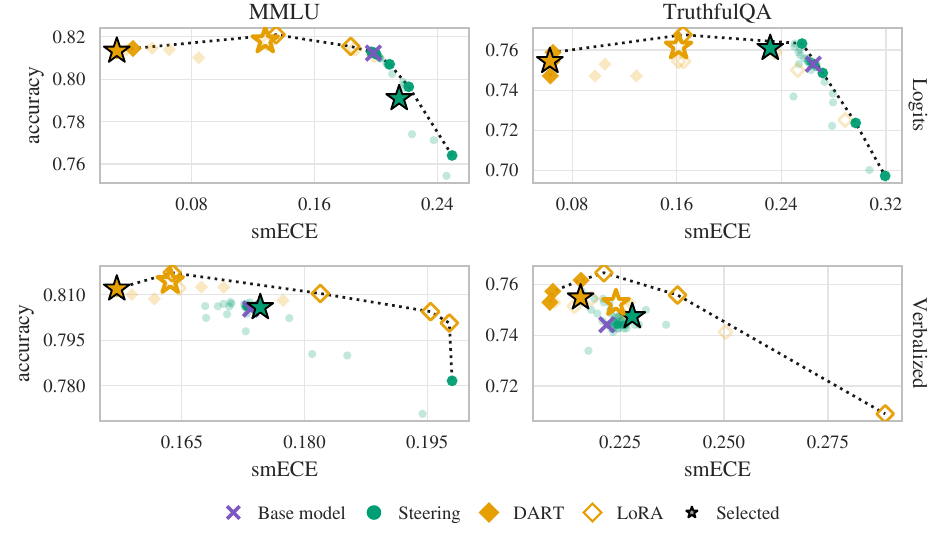}
    \caption{\textbf{Trade-off between calibration error and accuracy under the logit-based (top) and verbalized (bottom) confidence elicitation protocols on MMLU (left) and TruthfulQA (right)}. Results shown for \texttt{Llama-3.3-70B-Instruct}. Each point represents one hyperparameter configuration. Stars mark the selected configuration for each method. Lower smECE and higher accuracy values are better.}
    \label{fig:pareto_combined_smece_70B}
\end{figure*}

\Cref{fig:pareto_verbalized_euro_70B,fig:pareto_combined_smece_70B} show the trade-off between calibration and accuracy across hyperparameter configurations for \texttt{Llama-3.3-70B-Instruct}, using AUC-EURO and smECE for both confidence elicitation protocols. Under verbalized confidence, steering remains close to the base model in both datasets, while DART only improves AUC-EURO in MMLU. This further highlights that the relative performance of confidence elicitation protocols and the impact of our interventions, vary across models and tasks~\citep{tian-etal-2023-just,shrivastava2023llamasknowgptsdont}.

\FloatBarrier
\section{Results with Post-hoc Calibration}
\label{app:cross-model-results-ts}
\begin{figure*}[]
    \centering
    \includegraphics[width=\textwidth]{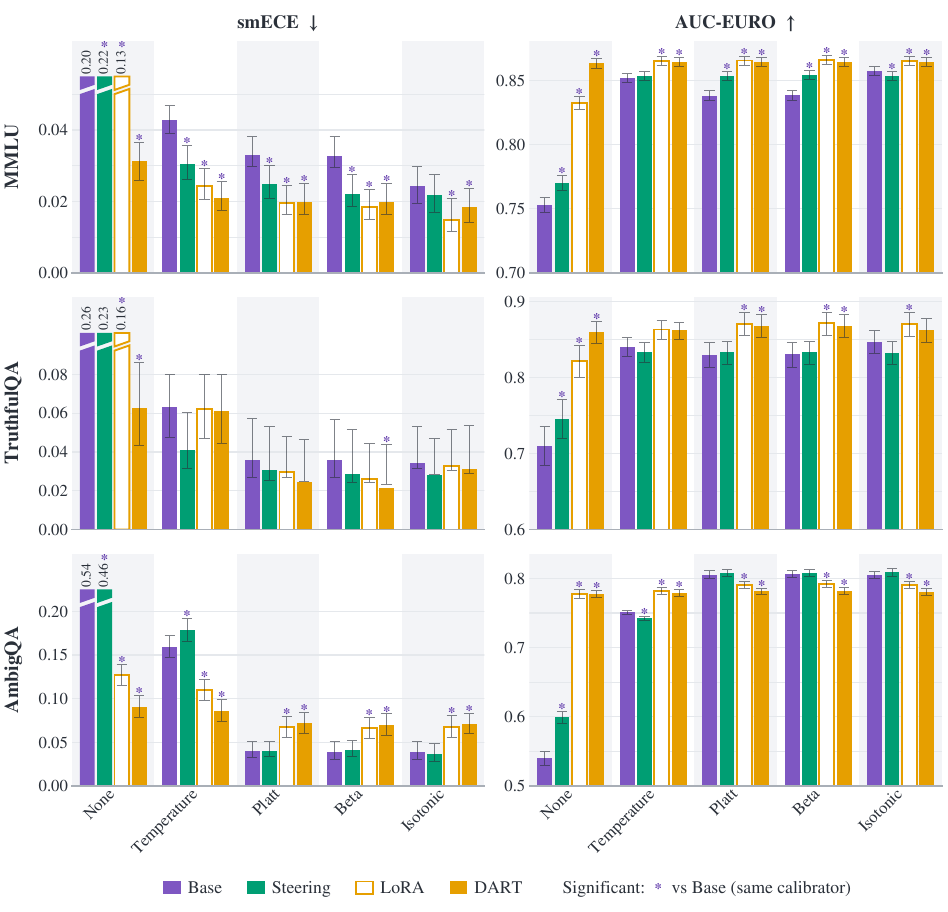}
    \caption{\textbf{Post-hoc calibration for} \texttt{Llama-3.3-70B-Instruct} under the \textbf{logits} protocol.
    smECE and AUC-EURO are shown for the base model, steering, LoRA, and DART, as trained (None) and after temperature scaling, Platt scaling, beta calibration, and isotonic regression, on MMLU (top), TruthfulQA (middle), and AmbigQA (bottom). The smECE bars for None are truncated, with their actual values labeled above. Error bars show 95\% bootstrap intervals. * marks a significant difference from the base model under the same calibrator.}
    \label{fig:post_hoc_logit_70B}
\end{figure*}

\begin{figure*}[]
    \centering
    \includegraphics[width=\textwidth]{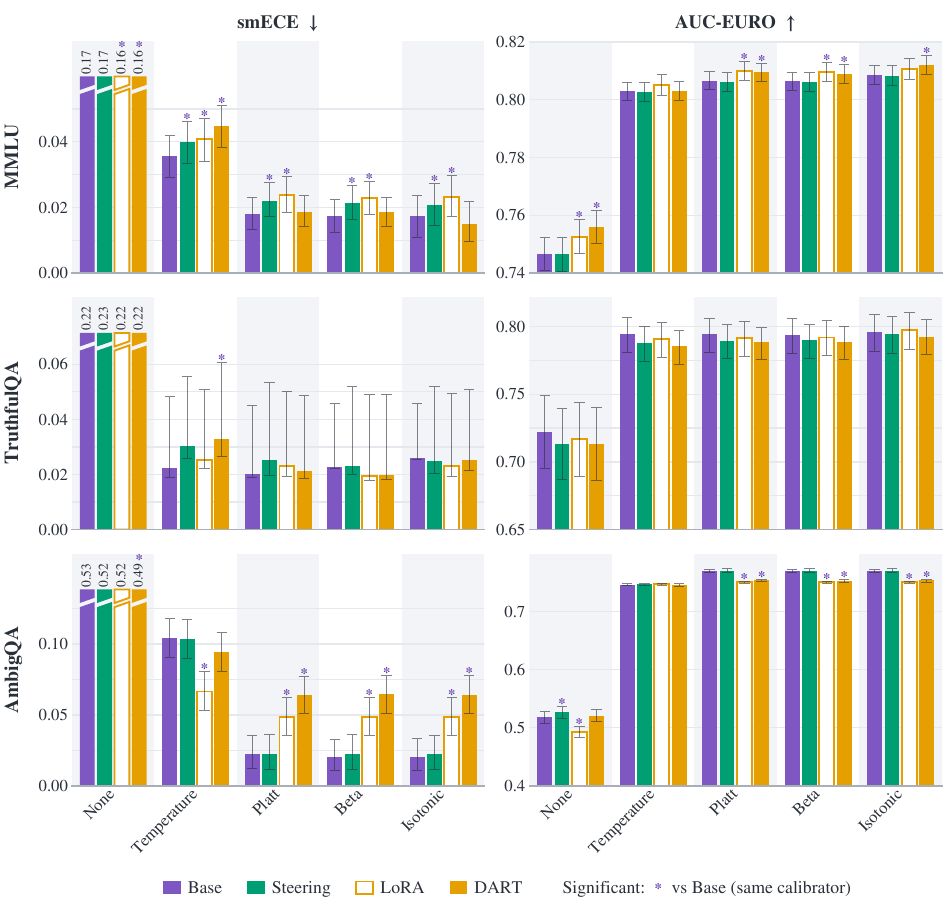}
    \caption{\textbf{Post-hoc calibration for} \texttt{Llama-3.3-70B-Instruct} under the \textbf{verbalized} protocol.
    smECE and AUC-EURO are shown for the base model, steering, LoRA, and DART, as trained (None) and after temperature scaling, Platt scaling, beta calibration, and isotonic regression, on MMLU (top), TruthfulQA (middle), and AmbigQA (bottom). The smECE bars for None are truncated, with their actual values labeled above. Error bars show 95\% bootstrap intervals. * marks a significant difference from the base model under the same calibrator.}
    \label{fig:post_hoc_verbalized_70B}
\end{figure*}

In~\cref{fig:post_hoc_logit_70B,fig:post_hoc_verbalized_70B} we report results for \texttt{Llama-3.3-70B-Instruct} across different tasks and confidence elicitation protocols, after applying four standard post-hoc calibration methods, namely temperature scaling, Platt scaling, beta calibration, and isotonic regression. Calibrators are fitted and evaluated following the procedures described in~App.~\ref{app:evaluation}. The other six models show the same pattern for LoRA and DART with logit confidence. The only exception is steering, whose AUC-EURO gain survives calibration only for \texttt{Llama-3.3-70B-Instruct}.

Temperature scaling rescales each model's output probabilities using a single scalar parameter fit on the validation set. Platt scaling fits a logistic mapping from the original score to a calibrated probability, while beta calibration uses a more flexible parametric mapping based on the log probabilities of the positive and negative classes. Isotonic regression instead learns a non-parametric monotonic mapping from confidence scores to calibrated probabilities. 

These methods operate only on the produced confidence estimates by correcting for over or under confidence without changing the model's predictions. Task performance is therefore unchanged by construction. By contrast, our interventions modify the model itself, potentially affecting both calibration and task accuracy. Importantly, post-hoc calibration methods are complementary to our proposed interventions and can be applied on top of steering or DART to further improve calibration, although the gains are not consistently additive.

\FloatBarrier
\section{Control Experiments}
\label{app:controls}
In this section, we study whether calibration improvements are specific to ambiguity or can also be achieved by steering along directions representing different concepts.

\subsection{Alternative Concept Controls}

The concept that we consider is \textit{sentiment} derived from CAD~\citep{kaushik2020learningdifferencemakesdifference}, applied at the best strength by EURO using the same steering procedure.
As shown in~\cref{fig:steering_controls}, steering along the control direction has only minimal effects on calibration compared to the ambiguity vector.

\subsection{Randomized Controls}
\begin{figure*}[]
    \centering
    \includegraphics[width=\textwidth]{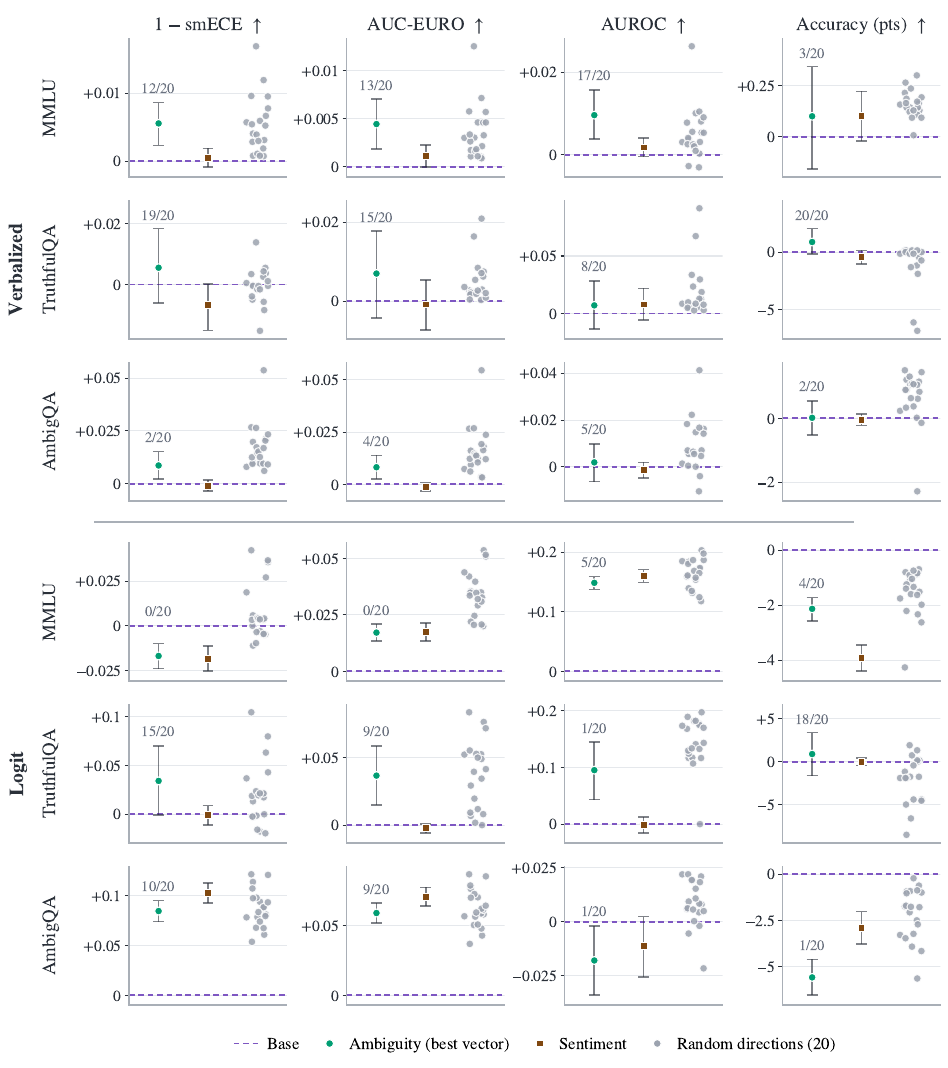}
    \caption{Comparison of \textbf{ambiguity steering with control directions} for \texttt{Llama-3.3-70B-Instruct} on MMLU, TruthfulQA and AmbigQA. We report the change from the base model in 1\,$-$\,smECE,
    AUC-EURO, AUROC and accuracy under steering with ambiguity, sentiment and 20
    random directions. Error bars are 95\% bootstrap intervals. The number
    above the ambiguity vector is how many of the 20 random directions it outperforms.}
    \label{fig:steering_controls}
\end{figure*}

\paragraph{Steering with random directions}
We additionally compare ambiguity steering against 20 random directions. As shown in~\cref{fig:steering_controls}, steering in general has minimal effects on calibration and they are mainly focused in verbalized confidence elicitation. While ambiguity steering outperforms most random directions under certain settings, the effect does not consistently hold across datasets and confidence elicitation protocols. This further highlights that the two protocols are not interchangeable, as the effects of steering depend on the task and evaluation setting. 

\begin{figure*}[]
    \centering
    \includegraphics[width=\textwidth]{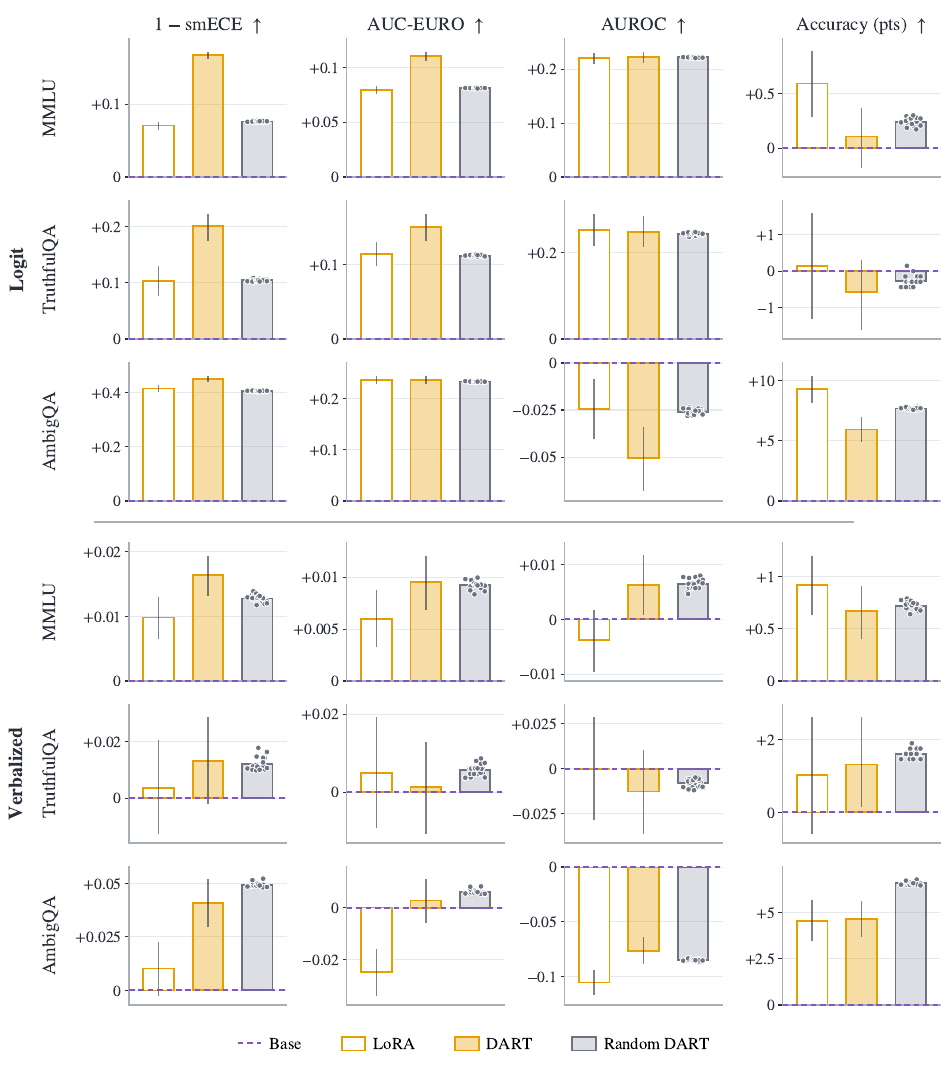}
    \caption{Random-span DART control for \texttt{Llama-3.3-70B-Instruct} on MMLU, TruthfulQA and AmbigQA. We report the change from the base model in 1\,$-$\,smECE, AUC-EURO, AUROC and accuracy under both confidence protocols. The gray bar is the mean of 20 random spans of the same rank. Error bars are 95\% bootstrap intervals.}
    \label{fig:random_dart}
\end{figure*}

\paragraph{Training LoRA with random directions}
To further study whether DART's substantial improvements in calibration depend on the ambiguity subspace, we replace it with a random subspace of the same dimensionality, while keeping the training procedure the same. ~\Cref{fig:random_dart} shows that training with random directions also improves calibration compared to the base model, however, DART consistently achieves lower smECE (higher 1\,$-$\,smECE) and higher AUC-EURO across all tasks and confidence elicitation protocols.

\FloatBarrier
\section{AI Use}
LLMs were used for writing plotting code, latex formatting, and answering technical questions through search engine's AI overview features.

\end{document}